\documentclass{article}

\usepackage{microtype}
\usepackage{graphicx}
\usepackage{subcaption}
\usepackage{booktabs} % for professional tables
\usepackage[most]{tcolorbox}
\usepackage{tabularx}
\usepackage{enumitem}
\usepackage{xcolor}

\usepackage{hyperref}
\usepackage{multirow}
\usepackage{subcaption}

\usepackage[preprint]{icml2026}

\usepackage{amsmath}
\usepackage{amssymb}
\usepackage{mathtools}
\usepackage{amsthm}
\usepackage{xspace}

\usepackage[capitalize,noabbrev]{cleveref}

\newtcolorbox{updateexample}[1]{
  enhanced,
  colback=gray!4,
  colframe=gray!40,
  boxrule=0.6pt,
  arc=2mm,
  left=2mm,right=2mm,top=1.5mm,bottom=1.5mm,
  title=\textbf{Example #1},
  fonttitle=\small,
}
\newtcolorbox{promptbox}[1][]{
  colback=gray!5, % Light gray background
  colframe=gray!50, % Darker gray frame
  fonttitle=\bfseries,
  coltitle=black,
  title={#1},
  boxrule=0.5mm,
  sharp corners,
  fontupper=\small\ttfamily, % Typewriter font for the content
  left=5pt, right=5pt, top=5pt, bottom=5pt
}

\theoremstyle{plain}

\theoremstyle{definition}

\theoremstyle{remark}

\renewcommand{\algorithmicrequire}{\textbf{Input:}}
\renewcommand{\algorithmicensure}{\textbf{Output:}}
\newcommand{\algorithmicparams}{\textbf{Parameters:}}
\newcommand{\PARAMS}{\item[\algorithmicparams]}

\usepackage[textsize=tiny]{todonotes}

\newcommand{\method}{\textsc{CoRSA}\xspace}

\newcommand{\Method}{\textbf{\underline{Co}}nflict-\textbf{\underline{R}}esolving and \textbf{\underline{S}}harpness-\textbf{\underline{A}}ware Minimization\xspace}

\usepackage{xcolor}
\definecolor{forestgreen}{HTML}{228B22}
\usepackage{booktabs}
\usepackage{tabularx}
\usepackage{pifont} % For checkmarks/crosses (optional)

\newcommand{\corr}[1]{\textcolor{green!60!black}{#1}}      % Correct text
\newcommand{\wrong}[1]{\textcolor{red}{#1}}        % Wrong text

\icmltitlerunning{Conflict-Resolving and Sharpness-Aware Minimization for Generalized Knowledge Editing with Multiple Updates}

\begin{document}

\twocolumn[
  \icmltitle{Conflict-Resolving and Sharpness-Aware Minimization for Generalized Knowledge Editing with Multiple Updates}



  \icmlsetsymbol{equal}{*}

  \begin{icmlauthorlist}
    \icmlauthor{Duy Nguyen}{yyy}
    \icmlauthor{Hanqi Xiao}{yyy}
    \icmlauthor{Archiki Prasad}{yyy}
    \icmlauthor{Elias Stengel-Eskin}{comp}
    \icmlauthor{Hyunji Lee}{yyy}
    \icmlauthor{Mohit Bansal}{yyy}
  \end{icmlauthorlist}

  \icmlaffiliation{yyy}{UNC Chapel Hill}
  \icmlaffiliation{comp}{The University of Texas at Austin}

  \icmlcorrespondingauthor{Duy Nguyen}{duykng@cs.unc.edu}

  \icmlkeywords{Machine Learning, ICML}

  \vskip 0.3in
]



\printAffiliationsAndNotice{}  % no special notice (required even if empty)

\begin{abstract}
Large language models~(LLMs) rely on internal knowledge to solve many downstream tasks, making it crucial to keep them up to date. Since full retraining is expensive, prior work has explored efficient alternatives such as model editing and parameter-efficient fine-tuning. However, these approaches often break down in practice due to poor generalization across inputs, limited stability, and knowledge conflict. To address these limitations, we propose the \method (\Method) training framework, a parameter-efficient, holistic approach for knowledge editing with multiple updates. \method{} tackles multiple challenges simultaneously: it \textit{improves generalization} to different input forms and \textit{enhances stability} across multiple updates by minimizing loss curvature, and \textit{resolves conflicts} by maximizing the margin between new and prior knowledge. Across three widely used fact editing benchmarks, \method{} achieves substantial gains in generalization, outperforming baselines with average absolute improvements of 12.42\% over LoRA and 10\% over model editing methods. With multiple updates, it maintains high update efficacy while reducing catastrophic forgetting by 27.82\% compared to LoRA. \method also generalizes to the code domain, outperforming the strongest baseline by 5.48\% Pass@5 in update efficacy. Our code is available at \url{https://github.com/duykhuongnguyen/CoRSA}. 
\end{abstract}

\section{Introduction} \label{sec:intro}
Given the importance of keeping large language models (LLMs) up-to-date and the high cost associated with retraining them, a growing body of work focuses on effectively updating the knowledge encoded in an LLM's parameters~\citep{ref:wang2024knowledge, ref:zhang2024comprehensive, ref:yao2023editing}. Across these lines of prior work, three key requirements for effective knowledge updating in LLMs have been identified. \textbf{First}, because LLMs must handle diverse input forms, updates should \textit{generalize} beyond the specific phrasing of the edited examples~\citep{ref:wang2024knowledge}, while preserving performance on the general model's capabilities. \textbf{Second}, the update mechanism should support \textit{multiple revisions}~\citep{ref:wang2024wise, ref:jiang2025neuron}. As knowledge evolves over time, models must repeatedly modify overlapping facts or behaviors while ensuring that prior updates can be revised or reverted without degrading unrelated updates. \textbf{Third}, since LLMs are pretrained on large corpora, they often encode prior knowledge that can \textit{conflict} with newly introduced information~\citep{ref:li2024unveiling, ref:xu2024knowledge}. This can result in reversion, where, despite an update, the model continues to produce the outdated information due to strong priors in its pre-trained weights~\citep{ref:xie2024adaptive, ref:bi2025decoding}. An effective knowledge editing method should thus minimize interference with the model’s existing knowledge~\citep{ref:ni2024forgetting, ref:li2025forget}. 

\begin{figure*}
    \centering
    \includegraphics[width=1.0\linewidth]{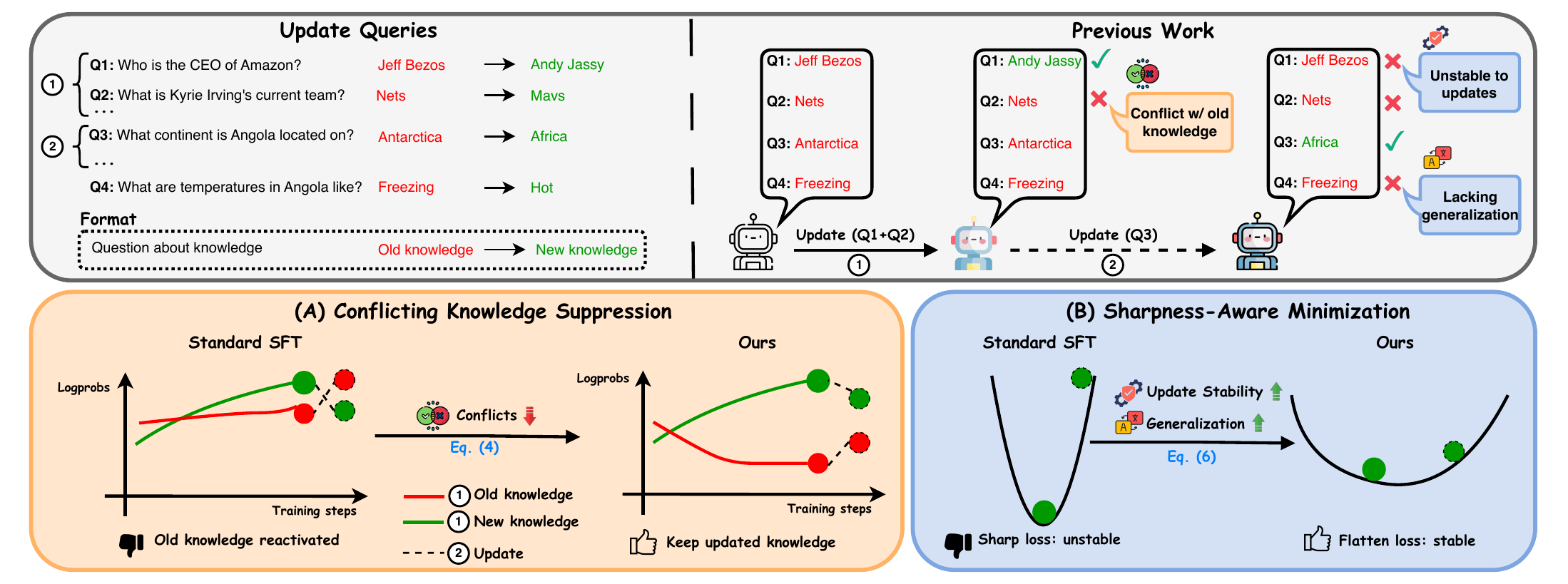}
    \caption{\textbf{Overview of \method{}.} \textbf{(Top) Limitations of previous work:} Existing approaches often fail to resolve conflicts, show poor generalization on varied inputs and instability under multiple updates (update 1, 2 in the figure), leading to catastrophic forgetting. \textbf{(Bottom) \method{}:} We address these limitations through two mechanisms: \textbf{(A) Conflicting Knowledge Suppression:} We explicitly suppress outdated information (red line), creating a distinct separation from the new target (green line). \textbf{(B) Sharpness-Aware Minimization:} We minimize the sharpness of the loss landscape, leading to better generalization and stability against future parameter updates (dashed line).}
    \label{fig:overall}
\end{figure*}
A variety of techniques have been proposed for effective knowledge updating, including model editing~\cite{ref:meng2022locating, ref:meng2022mass, ref:fang2025alphaedit} and fine-tuning~\citep{ref:yu2024melo, ref:gangadhar2024model}. However, no method successfully resolves all three aforementioned key requirements simultaneously (see~\cref{fig:overall}, Top). Model editing methods~\citep{ref:meng2022locating, ref:meng2022mass, ref:fang2025alphaedit} typically perform direct modifications to model weights. While effective for small-scale or single updates, these methods are often unstable under multiple or continual updates~\citep{ref:duan2025related, ref:thede2025wikibigedit}. Recent methods attempt to mitigate this via complex external memory routing~\citep{ref:wang2024wise, ref:li2025elder} or data replay~\citep{ref:fang2025hippocampal}, but these introduce significant architectural overhead and data dependencies. Moreover, they frequently fail to generalize across diverse input forms such as paraphrases in factual knowledge or syntactic variations in code~\citep{ref:li2024unveiling, ref:he2025knowledge}. Fine-tuning-based approaches, such as parameter-efficient fine-tuning (PEFT)~\citep{ref:hu2022lora, ref:han2024parameter}, are lightweight and effective for acquiring new knowledge. However, these methods often interfere with the model's prior knowledge, causing the model to continue to produce outdated information or otherwise produce incorrect text~\citep{ref:ni2024forgetting, ref:jung2025come}. Recent work has explored reducing such conflicts by selectively forgetting or suppressing outdated knowledge that can conflict with new updates~\citep{ref:ni2024forgetting, ref:li2025forget}. However, these methods are largely focused in single update scenarios, and we find that they don't extend to multiple knowledge updates, new knowledge injection, or generalization across diverse inputs.

To close this gap, we present \method{}, a \Method training framework for multi-update knowledge editing using LoRA~\citep{ref:hu2022lora}. As shown in~\cref{fig:overall} (Bottom), \method targets \textit{all} three requirements outlined above: stable update under multiple revisions, strong generalization across diverse inputs, and effective resolution of conflicts with the model's prior knowledge. First, we find that a LoRA adapter's capacity for generalization and stability to future updates are closely tied to the geometry of the loss landscape, with flatter regions leading to improved generalization and stability~(\cref{sec:effect-generalization}). Therefore, during LoRA training, we explicitly \textit{flatten the loss landscape} during training using Sharpness-Aware Minimization~\citep[SAM;][]{ref:foret2021sharpnessaware}. Second, we observe that standard supervised fine-tuning~(SFT) can inadvertently increase the likelihood of the old or outdated information due to knowledge conflict (see~\cref{fig:loss_sft}). To address this, we jointly minimize the negative log-likelihood and \textit{maximize the separation between new and old knowledge} via Direct Preference Optimization~\citep[DPO;][]{ref:rafailov2023direct}, thereby improving both stability under future updates and conflict resolution. Finally, when training with SFT and DPO objectives, the conflicting gradients between them lead to a suboptimal solution for SAM.  To resolve this, we employ gradient projection to stabilize the training dynamics (see~\cref{fig:loss_pcgrad}) and improve generalization compared to standard SAM. 

Empirically, we validate the effectiveness of \method{} on factual knowledge update, including standard knowledge update and continual revisions of knowledge. On factual benchmarks (CounterFact~\citep{ref:meng2022mass}, ZsRE~\citep{ref:levy2017zero}, MQuAKE~\citep{ref:zhong2025mquake}), our approach consistently outperforms baselines in update generality, including LoRA fine-tuning, model editing ~\citep[MEMIT;][]{ref:meng2022mass}, and forget-then-learn approaches ~\citep[F-Learning;][]{ref:ni2024forgetting}.
Using Qwen-3-4B-Instruct, \method{} improves generalization by 12.42\% compared to LoRA and by 9.99\% compared to MEMIT on average across three datasets. Similarly, we observe consistent performance gains with Llama-3.1-8B-Instruct. Moreover, we show that \method{} substantially improves update efficacy, outperforming the best performing baseline F-Learning by 2.95\% in the continual revision setting (updating a specific fact multiple times) and by 3.62\% in the knowledge injection setting (incorporating new information into LoRA). Simultaneously, \method{} achieves the lowest forgetting rates on unrelated knowledge (27.46\% and 23.55\% in continual revision and knowledge injection setting, respectively), substantially outperforming F-Learning (46.13\% and 29.13\%). Moreover, we show that our method generalizes to the code domain, where existing model editing methods are not directly applicable. On CodeUpdateArena~\citep{ref:liu2024codeupdatearena}, \method{} effectively updates code functionality while preserving general coding capabilities, surpassing F-Learning by 3.16\% Pass@1 and 5.48\% Pass@5 in update efficacy.

\section{Problem Formulation} \label{sec:problem}
In this section, we present the problem setup for knowledge updating in~\cref{sec:notations}, followed by the motivation for designing a generalizable and stable knowledge updating method in~\cref{sec:effect-generalization}.

\subsection{Problem Setup} \label{sec:notations}
A knowledge update aims to incorporate new information that deviates from the LLM's prior parametric knowledge, while preserving other existing knowledge that should remain unchanged.
Formally, we consider a pre-trained LLM $\theta$, which encodes outdated knowledge represented by a dataset $\mathcal{D}_{\text{old}}=\{(x_i, y^-_i)\}_{i=1}^{M}$, where $x_i$ is a query or context (e.g., \emph{``Who is the CEO of Amazon?"}) and $y_i^-$ denotes the corresponding response such as a multi-token fact (e.g., \emph{``Jeff Bezos''}). 
Given an update dataset $\mathcal{D}_{\text{new}}=\{(x_i, y^+_i)\}_{i=1}^{M}$, the goal is to learn a parameter update $\phi$ such that the resulting model $\theta^\prime = \theta + \phi$ predicts the new target $y_i^+$ instead of the outdated $y_i^-$. In our setting, we parameterize $\phi$ using a LoRA~\citep{ref:hu2022lora} adapter applied to the linear layers of the base model $\theta$ (details in~\cref{sec:app-exp-setting}). The formulation above describes a \textit{single} knowledge update. As knowledge evolves, in practice, knowledge updates are often performed \textit{multiple} times. 
We refer to this as multi-update knowledge editing, where the model performs continuous revision as shown in 
\cref{fig:overall}. In our setting, we perform these updates on the same set of LoRA parameters $\phi$.

\subsection{Analysis of Generalization and Stability for Knowledge Updating} \label{sec:effect-generalization}
In this section, we provide a theoretical analysis of the adapter $\phi$, specifically analyzing two properties: its \textit{generalization} at test time and its \textit{stability} when subjected to future updates. For clarity, we consider a single update at an input $x$, where the base model encodes old knowledge $y^-$ but is updated to reflect new knowledge $y^+$. Our formulation naturally extends to batch updates, in which multiple samples are updated simultaneously with a LoRA adapter. 

In knowledge updating, the objective is to successfully integrate new information without interfering with existing knowledge, ensuring that the model does not revert to outdated information. To quantify this, we measure the model's preference for the new versus the old knowledge using the log-likelihood margin of the LoRA adapter\footnote{This objective is also straightforward to formulate as a standard NLL loss function.} $\phi$: 
\begin{equation} \label{eq:margin}
    m(\phi) \triangleq \log p_{\theta,\phi}(y^+ \mid x) - \log p_{\theta,\phi}(y^- \mid x).
\end{equation}
We assume the base model $\theta$ encodes the prior knowledge $y^-$ at input $x$. Therefore, in the absence of an adapter ($\phi=0$), the model prefers the old target $y^-$ over the new target $y^+$ (i.e., $m(0) < 0$). After training LoRA for knowledge updating, we obtain LoRA parameters $\phi^\star$ such that:
\[
    m(\phi^\star) = \gamma > 0.
\]
The margin $\gamma$ describes how much the model prefers the new information to the old information.

\begin{figure*}[t]
    \centering
    \begin{subfigure}[b]{0.49\textwidth}
        \centering
        \includegraphics[width=\linewidth]{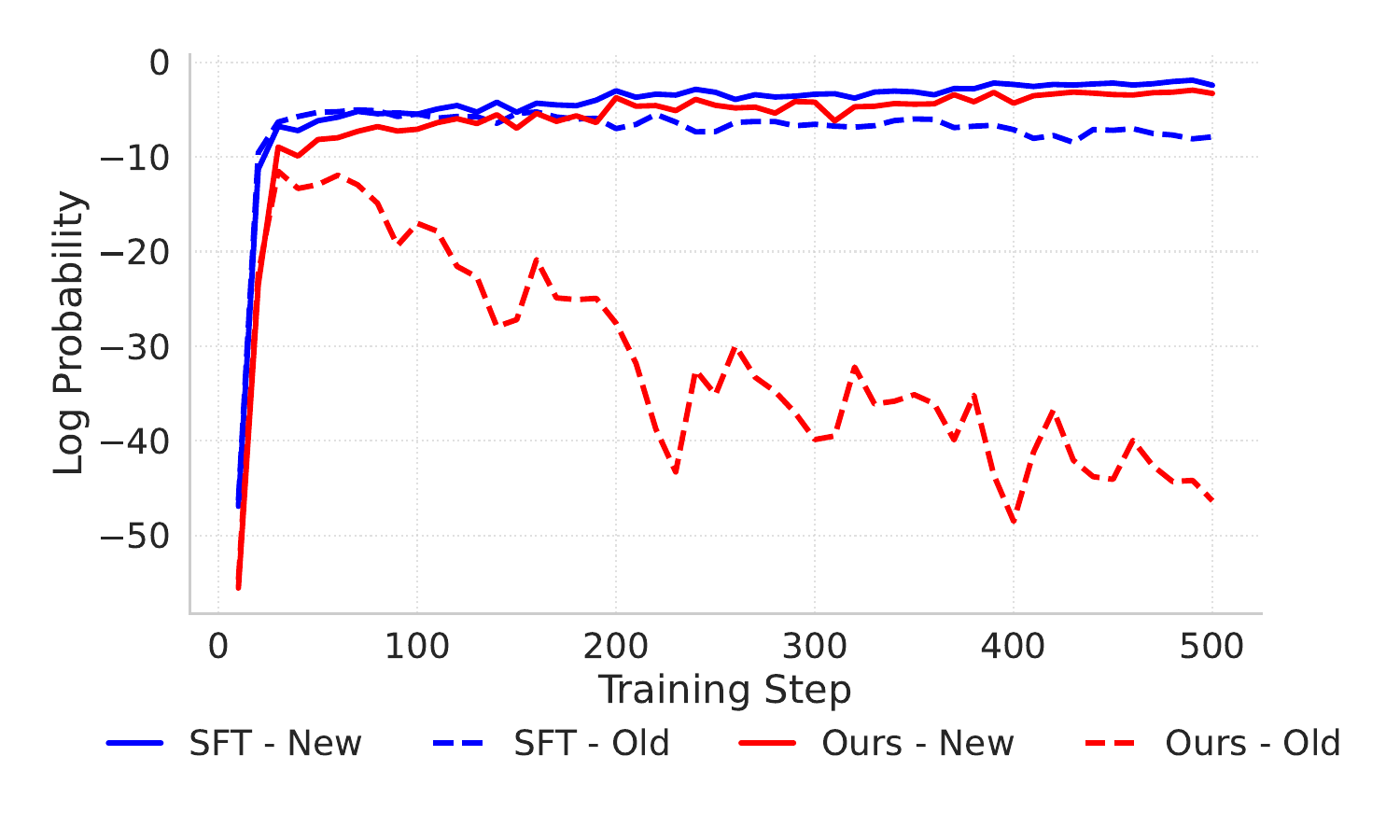}
        \caption{Standard SFT (blue) successfully learns the new knowledge but inadvertently maintains high probability for the old knowledge, leading to conflict. Ours (red) effectively separates the two by increasing $p(y_1 \mid x)$ while suppressing $p(y_0 \mid x)$ ($\gamma$ increases).}
        \label{fig:loss_sft}
    \end{subfigure}
    \hfill 
    \begin{subfigure}[b]{0.49\textwidth}
        \centering
        \includegraphics[width=\linewidth]{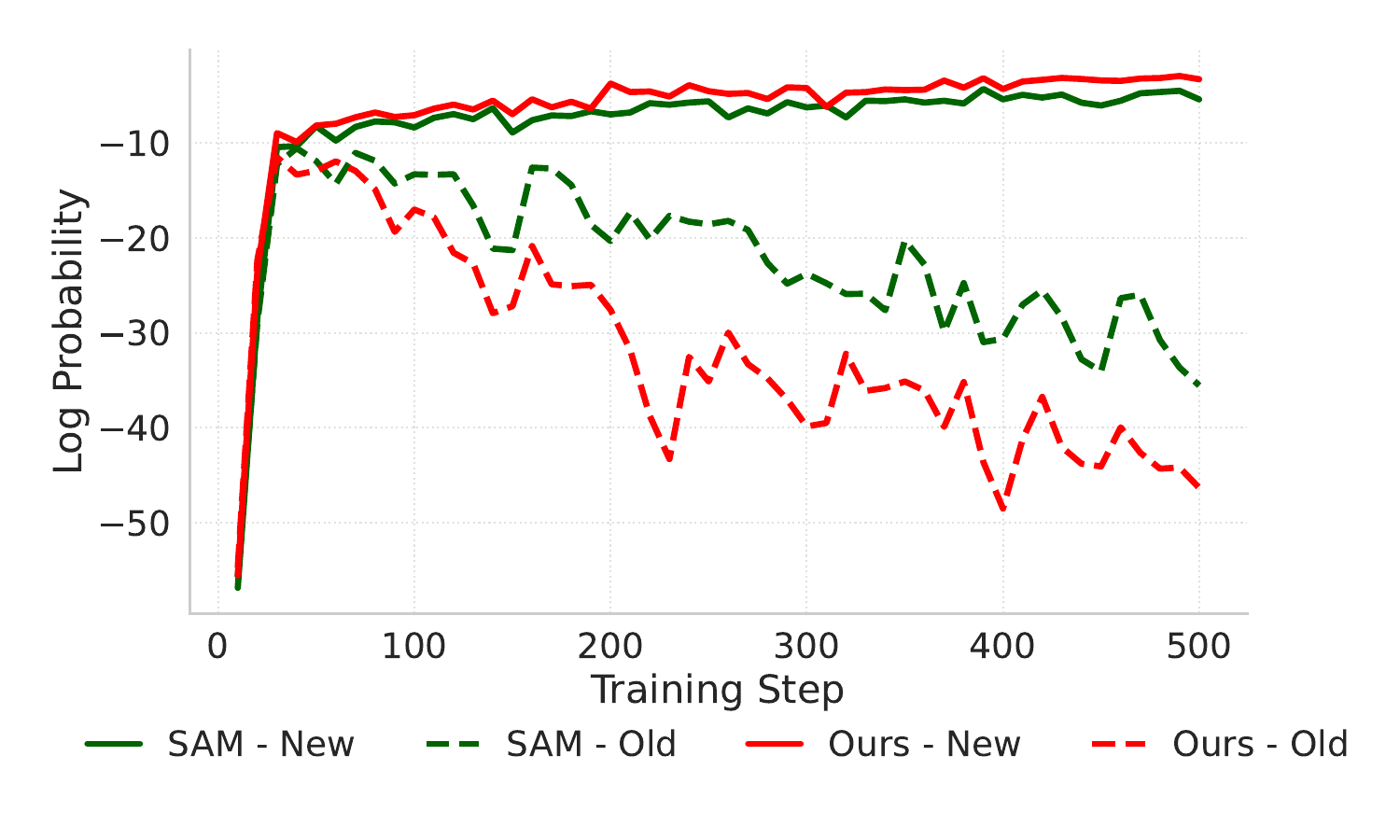}
        \caption{SAM without PCGrad (green) exhibits significant oscillations in log-probabilities, indicating destructive interference between objectives. In contrast, our method (red) employs PCGrad to resolve these conflicts, resulting in faster convergence.}
        \label{fig:loss_pcgrad}
    \end{subfigure}
    \caption{Log-probabilities for new knowledge (solid lines) and old knowledge (dashed lines) during training Qwen-3-4B on MQuAKE.}
    \label{fig:combined_loss}
\end{figure*}

\paragraph{Stability of LoRA Parameters.} As knowledge evolves, a knowledge update framework must support multiple updates of the adapter parameters $\phi$ (e.g., revising facts about the same entity, or inserting knowledge about new entities). We analyze the stability of a LoRA adapter under subsequent updates after the initial update $\phi^\star$. Additional training on new update data or objective produces a new adapter $\phi^\prime$. We model this change as an additive perturbation in parameter space $\phi^\prime=\phi^\star + \Delta$, where $\Delta$ denotes the parameter update induced by the subsequent updates. This perturbation is natural for LoRA because updates are implemented by directly optimizing the same low-rank parameters across time, and each new update continues gradient-based training from the current adapter state, so the next state differs from $\phi^\star$ by the accumulated optimizer steps (sum of gradients scaled by learning rates).

The future update $\Delta$ must not disrupt the knowledge encoded in $\phi^\star$. We define \textit{fallback} as any scenario where the margin reverts to $m(\phi^\star + \Delta) \leq 0$ under a future update $\Delta$. Assume $m(\phi)$ is twice differentiable in a neighborhood of $\phi^\star$ and its Hessian is bounded by a scalar $\kappa$, which controls the local curvature (or sharpness) of the landscape:
\begin{equation}\label{eq:hessian}
    \|\nabla^2 m(\tilde{\phi})\| \leq \kappa \quad \text{for } \tilde{\phi} \text{ near } \phi^\star.
\end{equation}
For any $\Delta$, a second-order Taylor expansion gives:
\[
m(\phi^\star + \Delta)
=
m(\phi^\star)
+
\langle \nabla m(\phi^\star), \Delta \rangle
+
\frac{1}{2} \Delta^\top \nabla^2 m(\tilde{\phi}) \Delta,
\]
for some $\tilde{\phi}$ on the line segment between $\phi^\star$ and $\phi^\star + \Delta$, where $\langle\cdot,\cdot\rangle$ is the parameter-space inner product and $\nabla,\nabla^2$ are w.r.t.\ $\phi$. Using the Hessian bound in Eq.~\eqref{eq:hessian} and substituting $m(\phi^\star)=\gamma$, we have:
\[
m(\phi^\star+\Delta)
\ge
\gamma
+
\langle \nabla m(\phi^\star), \Delta\rangle
-
\frac{\kappa}{2}\|\Delta\|_2^2.
\]
Therefore, a \textit{sufficient condition} to ensure \emph{no fallback} (i.e., $m(\phi^\star+\Delta) > 0$) for the future update $\Delta$ is:
\begin{equation}
\gamma > -\langle \nabla m(\phi^\star), \Delta\rangle + \frac{\kappa}{2}\|\Delta\|_2^2.
\label{eq:fallback-cond}
\end{equation}
Eq.~\eqref{eq:fallback-cond} describes the case when an adapter remains stable under future updates: the update margin $\gamma$ must dominate the first-order term that depends on the future update direction, and a second-order term that grows with the local curvature. As the direction of the future updates $\Delta$ are unknown given that the facts or data points for those updates are not available at the current update, we cannot optimize for the first-order term directly. Therefore, this equation suggests two concrete objectives for stability during knowledge updating (or learning $\phi^\star$): (i) \emph{increasing} the post-update margin $\gamma$, and (ii) \emph{decreasing} the margin curvature $\kappa$.

\paragraph{Loss Flatness and Generalization.} Beyond these benefits, minimizing $\kappa$ encourages the solution to settle in flatter regions of the loss landscape, which has been theoretically and empirically shown to enhance generalization for neural networks~\citep{ref:keskar2017on, ref:neyshabur2017exploring}. However, this connection is underexplored in the knowledge updating settings for LLMs. In this work, we aim to bridge this gap by integrating sharpness minimization into the knowledge updating framework and empirically showing improved generalization to input variations such as semantic paraphrases at test time in~\cref{sec:exp}.

\section{\method{}: \Method} \label{sec:corsa}
In this section, building on the analysis in~\cref{sec:effect-generalization}, we introduce \method{}, a LoRA training method for knowledge updating that targets the two terms in Eq.~\eqref{eq:fallback-cond} by: (i) \textit{increasing} the post-update margin $m(\phi^\star)=\gamma$, and (ii) \textit{reducing} local sharpness $\kappa$ with respect to the LoRA parameters $\phi$. Condition (i) ensures update stability and effective conflict resolution, whereas condition (ii) improves generalization and update stability.

\paragraph{Conflicting Knowledge Suppression.} For condition (i), optimizing the LoRA adapter with SFT alone is often insufficient. The new knowledge set $\mathcal{D}_{\text{new}}$ and the old knowledge set $\mathcal{D}_{\text{old}}$ (defined in~\cref{sec:notations}) are frequently semantically similar~\citep{ref:zhang2025resolving}. Thus, as shown in \cref{fig:loss_sft}, gradients induced by fitting $y^+$ can partially counteract the suppression of $y^-$ (i.e., re-activate the old behavior). To explicitly increase the margin and enforce separation between new and old knowledge, we optimize the LoRA parameters $\phi$ using a combined objective: a standard SFT loss for learning the updated targets and a DPO~\citep{ref:rafailov2023direct} loss between the old knowledge and new knowledge. Concretely, leveraging the old knowledge dataset $\mathcal{D}_{\text{old}}=\{(x_i, y^-_i)\}_{i=1}^{M}$ and the new knowledge dataset $\mathcal{D}_{\text{new}}=\{(x_i, y^+_i)\}_{i=1}^{M}$ defined in~\cref{sec:notations}, we construct a paired dataset $\mathcal{D}_{\text{pairs}}$ consisting of triplets $(x_i, y_i^{-}, y_i^{+})$. In this formulation, $y_i^{+}$ represents the target updated output, while $y_i^{-}$ denotes the outdated (or competing) response for the same input $x_i$. The DPO term encourages the model to assign higher likelihood to $y_i^{+}$ than to $y_i^{-}$, thereby directly increasing the margin (see~\cref{fig:overall}A).

More formally, the loss function is:
\begin{equation}
\label{eq:stage2_loss}
\mathcal{L}_{\text{Update}}(\phi)
=
\mathcal{L}_{\textsc{sft}}(\phi)
+
\lambda \, \mathcal{L}_{\textsc{dpo}}(\phi),
\end{equation}
where $\lambda \ge 0$ is a hyperparameter\footnote{We provide the hyperparameter analysis and details for $\lambda$ in~\cref{sec:app-implementation}.}.  

The individual loss terms are defined as:
\begin{align*}
\mathcal{L}_{\textsc{sft}}(\phi)
&=
- \mathbb{E}_{(x_i,y_i^{+})\sim\mathcal{D_{\text{new}}}}
\Big[
\log p_{\theta,\phi}(y_i^{+} \mid x_i)
\Big],
\\
\mathcal{L}_{\textsc{dpo}}(\phi)
&=
- \mathbb{E}_{(x_i,y_i^{-},y_i^{+})\sim\mathcal{D}_{\text{pairs}}}
\Bigg[
\log \sigma \Bigg(
\beta \log
\frac{p_{\theta,\phi}(y_i^{+} \mid x_i)}
     {p_{\theta}(y_i^{+} \mid x_i)}
\nonumber\\
&\hspace{3.5em}
-
\beta \log
\frac{p_{\theta,\phi}(y_i^{-} \mid x_i)}
     {p_{\theta}(y_i^{-} \mid x_i)}
\Bigg)
\Bigg],
\end{align*}
with inverse temperature $\beta > 0$ and reference model $p_{\theta}$.

\paragraph{Sharpness-Aware Minimization.}
For condition (ii), to reduce sharpness of the objective with respect to the LoRA parameters, we optimize $\mathcal{L}_{\text{Update}}$ using Sharpness-Aware Minimization~\citep[SAM][]{ref:foret2021sharpnessaware}.
We define the SAM perturbation for LoRA parameters as:
\begin{equation}\label{eq:sam_eps}
\epsilon^\star(\phi)
=
\arg\max_{\|\epsilon\|_2\le \rho}\ \mathcal{L}_{\text{Update}}(\phi+\epsilon),
\end{equation}
where $\rho>0$ controls the neighborhood radius.
SAM then minimizes the adversarially-perturbed loss:
\begin{equation}\label{eq:sam_obj}
\min_{\phi}\ \mathcal{L}_{\text{SAM}}(\phi)
\;\triangleq\;
\mathcal{L}_{\text{Update}}\big(\phi+\epsilon^\star(\phi)\big).
\end{equation}
By optimizing against this, we force the model to find a flatter region that is robust to parameter shifts (see~\cref{fig:overall}B). Following~\citet{ref:foret2021sharpnessaware}, we approximate Eq.~\eqref{eq:sam_eps} with a single ascent step to identify the worst-case perturbation: 
\begin{equation}\label{eq:sam_approx}
\epsilon^\star(\phi)
=
\rho\,
\frac{\nabla_\phi \mathcal{L}_{\text{Update}}(\phi)}
{\left\|\nabla_\phi \mathcal{L}_{\text{Update}}(\phi)\right\|_2+\varepsilon},
\end{equation}
where $\varepsilon>0$ is a small constant for numerical stability.
We then update $\phi$ using the gradient $\nabla_\phi \mathcal{L}_{\text{Update}}(\phi+\epsilon^\star(\phi))$. While SAM directly minimizes the sharpness of the training loss $\mathcal{L}_{\text{Update}}$, this objective serves as a principled proxy for stabilizing the margin $m(\phi)$. In particular, the DPO term is a composition of the log-probability margin and a monotonic function (e.g., negative log-sigmoid). By the chain rule, the Hessian of the loss $\nabla^2\mathcal{L_{\textsc{dpo}}}$ includes a term proportional to $\nabla^2 m$. Thus, minimizing the sharpness of the loss $\mathcal{L}_{\text{Update}}$ via SAM implicitly regularizes the DPO margin and encourages smaller margin curvature $\kappa$. We provide the formal derivation of this connection and further details regarding the motivation and details of SAM in~\cref{sec:sam}.

\paragraph{Gradient Conflict Resolution.}
Because $\mathcal{L}_{\textsc{sft}}$ and $\mathcal{L}_{\textsc{dpo}}$ can induce conflicting gradients~\citep{ref:hong2024orpo}, directly applying SAM to their weighted sum can be suboptimal~(\cref{fig:loss_pcgrad}). SAM involves \emph{two} gradient computations: an ascent direction to construct the perturbation $\epsilon$ in Eq.~\eqref{eq:sam_approx}, and a descent direction evaluated at the perturbed point in Eq.~\eqref{eq:sam_obj}.  For Eq.~\eqref{eq:sam_approx}, if $\epsilon$ is constructed from a dominant gradient, SAM explores sharpness mostly for that objective. For Eq.~\eqref{eq:sam_obj}, if the final update ignores conflicts, the descent step can still move in a direction that harms the other objective.

To mitigate this, we use PCGrad~\citep{ref:yu2020gradient} to form a conflict-reduced direction for optimizing the LoRA parameters $\phi$.
Let $g_{\textsc{sft}}=\nabla_\phi \mathcal{L}_{\textsc{sft}}(\phi)$ and $g_{\textsc{dpo}}=\nabla_\phi \mathcal{L}_{\textsc{dpo}}(\phi)$.
PCGrad projects each gradient to remove components that oppose the other when their inner product is negative:
\begin{equation*}\label{eq:pcgrad_proj}
\Pi(g_i; g_j)
=
\begin{cases} 
g_i - \dfrac{g_i^\top g_j}{\|g_j\|_2^2+\varepsilon}\, g_j,
& \text{if } g_i^\top g_j < 0,\\[6pt]
g_i, & \text{otherwise}.
\end{cases}
\end{equation*}
We then define a weighted conflict-reduced direction consistent with
$\mathcal{L}_{\text{Update}}=\mathcal{L}_{\textsc{sft}}+\lambda\mathcal{L}_{\textsc{dpo}}$:
\begin{equation*}\label{eq:pcgrad_sum}
g_{\textsc{pc}}(\phi)
=
\Pi(g_{\textsc{sft}}; g_{\textsc{dpo}})
+
\lambda\,\Pi(g_{\textsc{dpo}}; g_{\textsc{sft}}).
\end{equation*}
We apply PCGrad in both stages of SAM so that both the neighborhood explored by SAM and the update direction are conflict-reduced. Concretely, we first construct the SAM perturbation using the PCGrad-merged direction:
\begin{equation*}\label{eq:pc_sam_eps}
\epsilon^\star(\phi)
=
\rho\,
\frac{g_{\textsc{pc}}(\phi)}
{\|g_{\textsc{pc}}(\phi)\|_2+\varepsilon},
\end{equation*}
and then update $\phi$ after applying PCGrad to gradients from the perturbed point:
\begin{equation*}\label{eq:pc_sam_update}
\phi
\leftarrow
\phi - \eta\,
g_{\textsc{pc}}\big(\phi+\epsilon^\star(\phi)\big),
\end{equation*}
where $\eta$ is the learning rate.
This coupling ensures that the sharpness-aware step is taken in a direction that jointly respects the update objective (SFT) and the preference-separation objective (DPO), while reducing gradient interference. We provide the detailed training algorithm for \method{} in~\cref{alg:method_pc_sam},~\cref{sec:app-algo}. Furthermore, in~\cref{sec:app-ablation}, we provide a detailed ablation study on the individual components of \method{}, showing the importance of each component to the overall results. 

\section{Results} \label{sec:exp}
We design experiments to test the requirements for an effective knowledge updating framework using factual knowledge update benchmarks. In~\cref{sec:exp-setting}, we first present the experimental setup. We then present the evaluation of generalization to varied input forms (\cref{sec:exp1-factual}), the revision of knowledge (\cref{sec:exp2-factual}), and the interference between old and new knowledge (\cref{sec:exp3-factual}).
\begin{table*}[ht]
\centering
\small
\caption{Comparison of generality and specificity between different methods across three standard factual knowledge updating benchmarks. The best results among baselines and our method are highlighted in \textbf{bold}, and the second best are \underline{underlined}.}
\resizebox{\textwidth}{!}{
\begin{tabular}{lcccccccccccc}
\toprule
& \multicolumn{6}{c}{\textbf{Llama-3.1-8B-Instruct}} & \multicolumn{6}{c}{\textbf{Qwen-3-4B-Instruct}} \\
\cmidrule(lr){2-7} \cmidrule(lr){8-13}
& \multicolumn{2}{c}{CounterFact} & \multicolumn{2}{c}{ZsRE} & \multicolumn{2}{c}{MQuAKE}
& \multicolumn{2}{c}{CounterFact} & \multicolumn{2}{c}{ZsRE} & \multicolumn{2}{c}{MQuAKE} \\
\cmidrule(lr){2-3} \cmidrule(lr){4-5} \cmidrule(lr){6-7}
\cmidrule(lr){8-9} \cmidrule(lr){10-11} \cmidrule(lr){12-13}
\textbf{Method} & Gen $\uparrow$ & Spec $\uparrow$ & Gen $\uparrow$ & Spec $\uparrow$ & Gen $\uparrow$ & Spec $\uparrow$ & Gen $\uparrow$ & Spec $\uparrow$ & Gen $\uparrow$ & Spec $\uparrow$ & Gen $\uparrow$ & Spec $\uparrow$ \\
\midrule
Base Model
& - & 70.28 
& - & 70.28 
& - & 70.28 
& - & 71.74 
& - & 71.74 
& - & 71.74 \\
\midrule
LoRA 
& 49.25 & \textbf{69.31} 
& 70.40 & \textbf{70.02} 
& 90.97 & \underline{69.89} 
& 25.29 & \underline{70.47} 
& 68.95 & \underline{71.46} 
& 47.47 & \textbf{71.51} \\

MEMIT 
& 46.95 & 59.55 
& 70.95 & 61.64 
& 90.80 & 58.53 
& 26.95 & 55.13 
& 69.10 & 58.75 
& 52.95 & 56.82 \\

F-Learning 
& \underline{51.85} & 66.85 
& \underline{71.30} & 67.32 
& \underline{95.20} & 65.04 
& \underline{28.71} & 58.29 
& \underline{70.00} & 60.17 
& \underline{58.70} & 59.42 \\

\method{} 
& \textbf{66.27} & \underline{68.92} 
& \textbf{72.60} & \underline{69.88} 
& \textbf{96.53} & \textbf{70.17} 
& \textbf{39.71} & \textbf{70.81} 
& \textbf{72.10} & \textbf{71.51}  
& \textbf{67.17} & \underline{71.28} \\
\bottomrule
\end{tabular}
}
\label{tab:fact-editing}
\end{table*}

\subsection{Experimental Setup} \label{sec:exp-setting}

\paragraph{Datasets.} We use three standard benchmarks for factual knowledge updating, including \textbf{CounterFact}~\citep{ref:meng2022locating}, a dataset designed to distinguish between memorization and deep knowledge updates; \textbf{ZsRE}~\citep{ref:levy2017zero}, a QA dataset where relations are defined by natural language; and \textbf{MQuAKE-Remastered}~\citep{ref:zhong2025mquake}, a multi-hop QA dataset that measures how well models propagate factual updates across chains of linked facts.

\paragraph{Baselines.} We compare our approach against several baseline categories, including LoRA (fine-tuning with SFT on new knowledge), model editing (MEMIT~\citep{ref:meng2022mass}), and forget-then-learn approaches (F-Learning\footnote{We compare to single LoRA baselines for fair comparison.}~\citep{ref:ni2024forgetting}). For LoRA-based methods, we set the rank to 32, alpha to 64, the learning rate to $2e-4$, and the effective batch size to 16. We provide more implementation details for each method in~\cref{sec:app-implementation}.

\paragraph{Models.} We use Llama-3.1-8B-Instruct~\citep{ref:grattafiori2024llama} and Qwen-3-4B-Instruct~\citep{ref:yang2025qwen3} as the base models. In~\cref{sec:app-scale}, we further show that our method scales effectively to larger models (Qwen3-14B).

\paragraph{Metrics.} Following standard knowledge editing settings~\citep{ref:meng2022locating}, we report the following metrics:
\begin{itemize}[noitemsep, topsep=0pt, leftmargin=*]
    \item \textbf{Generality:} The success rate on paraphrased or semantically similar prompts that express the same knowledge but with different wording.
    \item \textbf{Specificity:} Performance of an LLM's general capabilities on unrelated tasks or knowledge, used to evaluate whether the injection of new knowledge does not compromise broad language understanding. We use the widely adopted MMLU benchmark~\citep{ref:hendrycks2021measuring}.
\end{itemize}
Reliability (or Edit Success) metric~\citep{ref:meng2022mass} (accuracy on prompts used for updates) is uniformly high ($97\text{-}98\%$) across methods; we therefore report these results in~\cref{tab:fact-editing-full} and focus on generality, a more reliable metric for update quality~\citep{ref:cohen2024evaluating, ref:gupta2024model}. 

\subsection{Individual Factual Knowledge Updates} \label{sec:exp1-factual}
Here, we evaluate the efficacy of standard knowledge updating and its impact on the model's broader capabilities. Specifically, we perform a single-batch update using the full dataset for each dataset and measure both the generality and specificity for every baseline.

\paragraph{Results.}~\cref{tab:fact-editing} demonstrates that \method{} consistently achieves the highest generalization across all datasets and models. On the Qwen-3-4B-Instruct model, we observe a substantial improvement on the CounterFact dataset, where our method \emph{outperforms} F-Learning by 11.00\% and LoRA by 14.42\%. This \emph{extends to the Llama-3.1-8B-Instruct model}. Similarly, on MQuAKE with Qwen-3-4B-Instruct, we outperform F-Learning by 14.22\% and LoRA by 19.70\%. We provide examples showing generalization to diverse input forms in~\cref{sec:app-examples} (\cref{tab:counterfact-examples} and~\cref{tab:mquake-examples}). 

Moreover, \method{} demonstrates \emph{superior stability} than F-Learning on MMLU. On CounterFact with Qwen-3-4B-Instruct, we maintain a Specificity of 70.81\%, which is closer to the base model's 71.74\% than F-Learning, which drops to 58.29\%. Across datasets, the performance drop of our method is marginal ($\sim$0--1.3\%), showing that \method{} effectively preserves the model's general capabilities.

\subsection{Continual Knowledge Revision} \label{sec:exp2-factual}
In this section, we conduct the experiments to test the stability of methods under multiple updates. In the context of LoRA-based approaches, the adapter trained in the previous section serves as a persistent module for storing knowledge. When specific entries are updated, the model must achieve high update efficacy on the new information while effectively mitigating catastrophic forgetting of previously learned information. We define continual \textbf{Update Efficacy} as the model's generalization performance on the updated knowledge, and \textbf{Forgetting} as the degradation in performance on the knowledge previously stored in the adapter. In this subsection, we evaluate the stability of our adapter under two distinct continual update settings.  

\paragraph{Cross-Dataset Knowledge Injection.} First, we evaluate the performance of the adapter when injecting entirely new, distinct knowledge into an already trained module. The goal of this experiment is to achieve high performance on the new dataset (high update efficacy) while maintaining the accuracy of the knowledge originally stored in the adapter (low forgetting). To test this, we take the adapter trained on the CounterFact dataset in the previous section and continue training it on the MQuAKE dataset. We then evaluate the model's performance on the new MQuAKE entries and measure the forgetting rate on the original CounterFact evaluation set. This setting tests the adapter's capacity to serve as a cumulative knowledge store across data sources.~\cref{tab:cross-inject} demonstrates that \method{} outperforms F-Learning in forgetting by 5.58\% and 11.28\% in Llama-3.1-8B and Qwen-3-4B, respectively. These results demonstrate that our objective allows for \emph{high update efficacy} in new domains \emph{without overwriting} the source domain knowledge. Conversely, \emph{standard LoRA is notably unstable} in this setting with the higher rate of catastrophic forgetting compared to our method (e.g., 12.19\% on Llama-3.1-8B).
\begin{table}[t]
\centering
\small
\caption{Trade-off between update efficacy and forgetting when adapting a CounterFact-trained adapter to the MQuAKE dataset.}
\label{tab:cross-inject}
\begin{tabular}{lcccc}
\toprule
& \multicolumn{2}{c}{\textbf{Llama-3.1-8B}} & \multicolumn{2}{c}{\textbf{Qwen-3-4B}} \\
\cmidrule(lr){2-3} \cmidrule(lr){4-5}
\textbf{Method} & Update $\uparrow$ & Forget $\downarrow$ & Update $\uparrow$ & Forget $\downarrow$ \\
\midrule
LoRA        & 87.43 & 35.74 & 44.87 & \underline{20.72} \\
MEMIT       & 85.92 & 52.86 & 41.10 &  34.17            \\
F-Learning  & \underline{90.37} & \underline{29.13} & \underline{48.96} & 25.39 \\
\method{}   & \textbf{90.87} & \textbf{23.55} & \textbf{54.77} & \textbf{14.11} \\
\bottomrule
\end{tabular}
\end{table}

\begin{table}[t]
\centering
\small
\caption{Trade-off between learning new temporal facts and retaining non-updated historical information on AToKe dataset.}
\label{tab:temporal-update}
\begin{tabular}{lcccc}
\toprule
& \multicolumn{2}{c}{\textbf{Llama-3.1-8B}} & \multicolumn{2}{c}{\textbf{Qwen-3-4B}} \\
\cmidrule(lr){2-3} \cmidrule(lr){4-5}
\textbf{Method} & Update $\uparrow$ & Forget $\downarrow$ & Update $\uparrow$ & Forget $\downarrow$ \\
\midrule
LoRA        & \underline{97.47} & 55.28 & \textbf{91.59} & 53.39 \\
MEMIT       & 89.29 & 61.38 & 82.21 & 67.48 \\
F-Learning  & 95.09 & \underline{46.13} & 87.61 & \underline{37.32} \\
\method{}   & \textbf{98.04} & \textbf{27.46} & \underline{91.23} & \textbf{21.15} \\
\bottomrule
\end{tabular}
\end{table}

\paragraph{Temporal Knowledge Updates.} Second, we evaluate the performance of the adapter when the model is updated with a stream of knowledge changes over time. The objective is to update specific knowledge subsets (high update efficacy) while preserving the integrity of previously learned information in the adapter that is not subject to the current update (low forgetting). For this experiment, we use the AToKe-ME dataset~\citep{ref:yin2024history}, which is specifically designed to evaluate how models handle \textit{multiple} factual knowledge updates over time in the real world with temporal fact chains (e.g., the progression of Amazon CEOs in~\cref{fig:overall}). As shown in~\cref{tab:temporal-update}, \method{} achieves the \emph{highest update efficacy} (98.04\%) while simultaneously maintaining the \emph{lowest forgetting rate} (27.46\%) on Llama-3.1-8B. In contrast, while standard LoRA and F-Learning adapt well to new information, they \emph{suffer from catastrophic forgetting} with substantial forgetting rates of 55.28\% and 46.13\%, respectively. In~\cref{sec:app-samples-forgetting}, we also show that the increase of catastrophic forgetting is directly correlated with the total number of samples used for updates. 

\begin{figure*}
    \centering
    \includegraphics[width=0.9\linewidth]{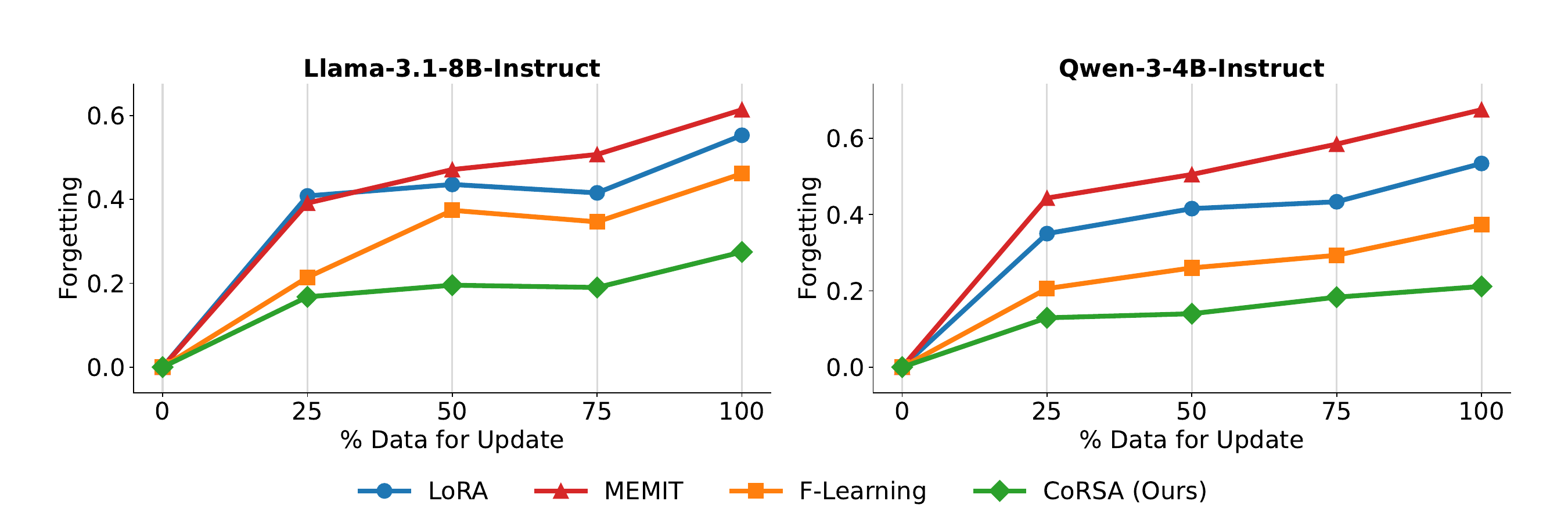}
    \caption{Trade-offs between the percentage of data used for updates and forgetting in the continual knowledge revision setting. \method{} consistently demonstrates superior stability, achieving substantially lower forgetting rates compared to baselines across all data settings.}
    \label{fig:forgetting-vs-update}
\end{figure*}

\subsection{Interference with Old Knowledge} \label{sec:exp3-factual}
Failure cases in updating are often driven by a strong prior on old information, resulting in conflicting, older knowledge being recruited or ambiguous answers being produced. In this section, we analyze whether the model truly overwrites the prior belief or leaves the outdated information active after updating knowledge.

\paragraph{Retention of Outdated Knowledge with Individual Factual Knowledge Updates.}
\begin{table}[t]
\centering
\small
\caption{Outdated Knowledge Retention after a single update using Qwen-3-4B-Instruct.}
\label{tab:okr-single}
\begin{tabular}{lccc}
\toprule
& \multicolumn{3}{c}{\textbf{Outdated Knowledge Retention} $\downarrow$} \\
\cmidrule(lr){2-4}
\textbf{Method}
& CounterFact
& ZsRE
& MQuAKE \\
\midrule
LoRA               & 9.81 & 6.80 & 3.57 \\
MEMIT              & 8.58 & 8.32 & 5.50 \\
F-Learning         & \underline{7.07} & \underline{5.57} & \underline{1.20} \\
\method{} & \textbf{1.28} & \textbf{2.21} & \textbf{0.03} \\
\bottomrule
\end{tabular}
\end{table}
After the single update in~\cref{sec:exp1-factual}, we run the model on the same paraphrased prompts used for testing Generality in~\cref{tab:fact-editing}, and define ``Old Knowledge Retention'' (\%) as the frequency with which the model still produces the outdated response given the query. As shown in~\cref{tab:okr-single}, \emph{baselines struggle to suppress prior beliefs}. For example, LoRA retains outdated knowledge in 9.81\% of cases on CounterFact and 6.80\% on ZsRE, explaining its lower generality score in \cref{tab:fact-editing}. In contrast, \method{} effectively \emph{mitigates this interference}, achieving substantially lower retention rates of 1.28\% and 2.21\% respectively, with near-zero retention (0.03\%) on MQuAKE. 

\begin{table}[t]
\centering
\small
\caption{Old Knowledge Activation after continual updates.}
\label{tab:oka-continual}
\begin{tabular}{lcc}
\toprule
& \multicolumn{2}{c}{\textbf{Old Knowledge Activation} $\downarrow$} \\
\cmidrule(lr){2-3}
\textbf{Method}
& Llama-3.1-8B 
& Qwen-3-4B \\
\midrule
LoRA               & 11.48 & 17.09 \\
MEMIT              & \underline{6.38} & 12.68 \\
F-Learning         & 8.29 & \underline{10.53} \\
\method{} & \textbf{3.71} & \textbf{7.72} \\
\bottomrule
\end{tabular}
\end{table}
\paragraph{Old Knowledge Reactivation with Continual Updates.} The failure to fully suppress outdated knowledge is critical in continual update settings. We observe a phenomenon of knowledge reactivation, where old facts that appeared to be successfully overwritten initially reactivate after the model is updated with new knowledge (see~\cref{tab:atoke_reactivation_examples} for examples). In this experiment, we define ``Old Knowledge Activation'' as the percentage of samples in which old knowledge is reactivated under the continual update setup in~\cref{tab:cross-inject}. In \cref{tab:oka-continual},  baselines show \emph{high susceptibility to this reversion}. For example, LoRA has an activation rate of 11.48\% on Llama-3.1-8B-Instruct and 17.09\% on Qwen-3-4B-Instruct. \method{} \emph{substantially reduces this risk}, achieving the lowest activation rates of 3.71\% and 7.72\%, respectively.

\section{Analysis}
In this section, we analyze the trade-offs between the number of samples and stability (\cref{sec:app-samples-forgetting}), evaluate the scalability to larger models (\cref{sec:app-scale}), and show the transferability of \method{} to the code domain (\cref{sec:exp-code}).
\subsection{Trade-off between Total Number of Samples and Stability} \label{sec:app-samples-forgetting}
In the continual knowledge revision setting, as the total number of training steps applied to a pretrained adapter increases (the total number of samples or updates increases), the existing knowledge learned in the adapter will also be forgotten more. In this section, we test this by varying the percentage of the total number of samples used for updates and see how baselines and our method perform. As shown in~\cref{fig:forgetting-vs-update}, the results show that forgetting increases for all methods as the percentage of data used for updates grows. However, the baseline methods, particularly MEMIT and LoRA, demonstrate a substantial rise in forgetting. In contrast, our \method{} demonstrates greater stability, maintaining the lowest level of forgetting across all data percentages for both Llama-3.1-8B-Instruct and Qwen-3-4B-Instruct models.

\begin{figure*}
    \centering
    \includegraphics[width=0.9\linewidth]{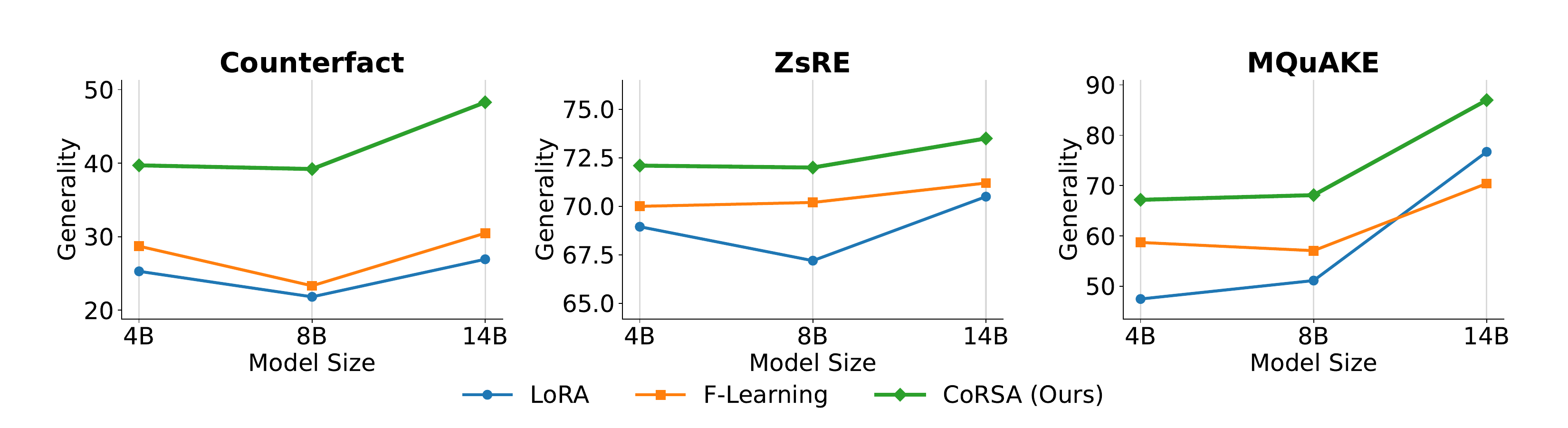}
    \caption{Generality of LoRA, F-Learning, and \method{} when updating Qwen models of different sizes (4B/8B/14B) on CounterFact, ZsRE, and MQuAKE. \method{} consistently yields the best generality across model sizes.}
\label{fig:qwen_scaling}
\end{figure*}

\subsection{Scalability to Larger Models} \label{sec:app-scale}
In this section, to assess the scalability of our method to larger models, we extend our evaluation to multiple sizes of the Qwen family (Qwen3-4B-Instruct, Qwen3-8B, and Qwen3-14B). For consistency, we use the identical experimental settings described in~\cref{sec:exp}. As shown in~\cref{fig:qwen_scaling}, \method{} consistently achieves higher generality than LoRA and F-Learning across all three model sizes and all datasets. For instance, the performance gap is the most substantial on CounterFact, where \method{} outperforms baselines by over 10\% across all scales (4B, 8B, and 14B). Even as the baselines improve with model size, they fail to close the gap, demonstrating that our training objective provides fundamental benefits to update stability.

\subsection{Transfer to Code Domain} \label{sec:exp-code}
\begin{table}[t]
\small
\centering
\caption{Comparison of UPass and SPass between different methods on CodeUpdateArena. Prepend has no SPass because it is parameter-free and behaves identically to the Base Model. Base Model has no UPass as it is not exposed to the update.}
\begin{tabular}{lcccc}
\toprule
 & \multicolumn{2}{c}{\textbf{UPass (Efficacy)} $\uparrow$} & \multicolumn{2}{c}{\textbf{SPass (Specificity)} $\uparrow$} \\
\cmidrule(lr){2-3} \cmidrule(lr){4-5}
\textbf{Method} & Pass@1 & Pass@5 & Pass@1 & Pass@5 \\
\midrule
Base Model
 & - & - 
 & 62.80 & 73.17 \\
\midrule
Prepend    
 & 15.37 & 22.11
 & - & - \\

LoRA     
 & \underline{17.68} & 27.78 
 & \underline{58.54} & \textbf{69.51} \\

F-Learning  
 & 17.26 & \underline{29.05} 
 & 56.71 & 67.07 \\

\method{}    
 & \textbf{20.42} & \textbf{34.53} 
 & \textbf{59.15} & \underline{68.90} \\
\bottomrule
\end{tabular}
\label{tab:code-editing}
\end{table}
In this section, we evaluate our method's ability to handle the structural and logical complexity of \emph{code} domain, a domain where previous locate-and-edit methods such as MEMIT are fundamentally inapplicable due to their reliance on encoding single factual vectors (discrete key-value pairs in the parameter space to map an old fact to a new one).

\paragraph{Setup.} We use \textbf{CodeUpdateArena}~\citep{ref:liu2024codeupdatearena}, a benchmark simulating API evolution. It consists of synthetic API updates (e.g., adding an argument `reverse=True' to a sort function) paired with program synthesis problems that require using the updated API. We compare our approach against Prepend (which adds the update description to the model context), LoRA, and F-Learning using Qwen-3-4b-Instruct. Following~\citet{ref:liu2024codeupdatearena}, we evaluate using \textbf{UPass (Efficacy)}, the pass rate on tests strictly requiring the new API, and \textbf{SPass (Specificity)}, the retained performance on unrelated HumanEval~\citep{ref:chen2021evaluating} tasks. For code update, we set the rank to 128, alpha to 256, and learning rate to $2e-5$ because code tasks typically involve complex dependencies and logic that require higher model capacity.

\paragraph{Results.}~\cref{tab:code-editing} shows that our method substantially outperforms LoRA and F-Learning by 2.74\% and 3.16\% in Pass@1 accuracy, respectively.  Importantly, this gain does not compromise the model's general coding abilities. On the specificity benchmark (SPass), our method retains more general coding capability, outperforming LoRA by 0.61\% and F-Learning by 2.44\% on Pass@1.

\section{Related Work}
\paragraph{Multiple Knowledge Updates.} LLMs often require updating methods that can be \emph{applied multiple times}~\citep{ref:dhingra2022time, ref:hartvigsen2023aging}. This property is closely related to continual learning~\citep{ref:kirkpatrick2017overcoming, ref:li2017learning}. However, unlike standard continual learning, which aims to mitigate forgetting across distinct downstream tasks, knowledge updating requires a more flexible paradigm where information for the same or different entities can be revised or reverted without degrading the general model's capabilities using lightweight techniques. This necessity motivates the use of modular parameter sets, such as FFNs or LoRA as containers for updates, enabling the retrieval, routing, and composition of modules~\citep{ref:huang2023lorahub, ref:wu2024mixture, ref:ostapenko2024towards}. For continual update, WISE~\citep{ref:wang2024wise} stores edits in a side FFN memory and use activation-based routing to switch between pre-trained and edited knowledge. Similarly, ELDER~\citep{ref:li2025elder} routes inputs through a Mixture-of-LoRA architecture, while MELO~\citep{ref:yu2024melo} directs inputs to neuron-indexed LoRA blocks via a vector database. Additionally, \citet{ref:fang2025hippocampal} explicitly replays past edits to maintain stability in sequential editing. In contrast to these methods, which \emph{rely on routing and external memory for retrieval or require past replay}, our method embeds the knowledge into LoRA and use sharpness-aware minimization to ensure that \emph{updates are stable}.

\paragraph{Generalizable and Domain-Transferable Updates.} A critical challenge in knowledge editing is generalization, ensuring that an update generalizes across various input forms. Benchmarks like RippleEdits~\citep{ref:cohen2024evaluating} and EVOKE~\citep{ref:zhang2025uncovering} are specifically designed to test this capability. Previous work such as~\citet{ref:mitchell2022fast} and~\citet{ref:de2021editing} address this by using hypernetworks to predict weight updates based on gradients to prevent overfitting to specific syntactic patterns. More recent work such as~\citep{ref:wei2025setke} uses bipartite matching to update sets of overlapping triplets simultaneously. Distinct from this line of work, which \emph{rely on complex auxiliary networks or specialized matching algorithms using chains of facts}, our approach achieves \emph{generalization directly through training} by explicitly optimizing for flat minima in the loss.

Another crucial property in knowledge update for LLMs is domain transferability, as LLMs are expected to work on diverse domains. While factual editing typically targets isolated associations, domains such as code require updating complex reasoning while preserving syntactic dependencies. While benchmarks for this setting have been introduced~\citep{ref:liu2024codeupdatearena, ref:misra2025gitchameleon}, existing model editing methods such as AlphaEdit~\citep{ref:fang2025alphaedit} \emph{struggle to generalize outside of fact-based tasks}. We aim to address this with a generalizable objective applied to LoRA, enabling \emph{effective updates across both factual and code domains}. 

\section{Conclusion}
In this work, we proposed \method{}, a holistic framework that targets three core requirements for knowledge updating, using complementary approaches: generalization to various inputs and stability to future updates by minimizing loss curvature with SAM, and mitigate interference by maximizing the margin between new and prior knowledge with DPO. 
Experiments on factual knowledge updating datasets demonstrate that \method{} outperforms strong baselines in generalization, substantially reducing catastrophic forgetting during continual revision and minimizing the retention of outdated knowledge. Furthermore, our analysis shows that \method{} maintains stability scales efficiently to larger models and transfers effectively to the code domain.

\section*{Acknowledgments}
This work was supported by NSF-AI Engage Institute DRL2112635, NSF-CAREER Award 1846185, DARPA ECOLE Program No. HR00112390060, Capital One Research Award, and an Apple PhD Fellowship. The views contained in this article are those of the authors and not of the funding agency.

\section*{Impact Statement}
The primary goal of this work is to improve the reliability of LLMs over time through efficient knowledge updates, enabling the update of new information without the substantial computational cost of re-training. 
However, we acknowledge that current updating methods, including ours, are imperfect and may introduce failures and unintended side effects, such as hallucinations. 
Moreover, while these techniques allow for updating the model with new knowledge, they could be exploited by malicious actors to inject misinformation, and are limited by the quality of their input data.
We believe that improving the generalization, stability, and reducing conflicts of updates, as proposed in this work is a necessary step toward mitigating these unintended side effects and reducing the risks in the deployment of LLMs.

\bibliography{bibliography}
\bibliographystyle{icml2026}

% APPENDIX

\newpage

\appendix
\onecolumn
\crefalias{section}{appendix}
\crefalias{subsection}{appendix}      

\section{Experimental Settings} \label{sec:app-exp-setting}
\subsection{Datasets}
In this subsection, we describe the datasets and explain how we preprocess them for the experiments. Each dataset provides a distinct set of updates for either factual or code domain. We report the dataset statistics in~\cref{tab:dataset-stats}.
\paragraph{Factual Knowledge Datasets}
\begin{itemize}
    \item \textbf{CounterFact}~\citep{ref:meng2022locating}: A  dataset containing facts designed to distinguish between memorization and deep knowledge updates for LLMs.
    \item \textbf{ZsRE}~\citep{ref:levy2017zero}: A question-answering dataset where relations are defined by natural language questions.
    \item \textbf{MQuAKE-Remastered}~\citep{ref:zhong2025mquake}: A dataset for reliable knowledge editing evaluation, measuring how well models propagate factual updates across linked facts.
    \item \textbf{AToKe-ME}~\citep{ref:yin2024history}: A temporal knowledge editing dataset designed to evaluate \emph{multiple sequential updates} of the same subject–relation pair over time. Each instance consists of a temporal fact chain in which the same fact is updated repeatedly across successive time periods. For the continual knowledge revision experiments in~\cref{sec:exp2-factual}, we select samples containing a total of three sequential updates. We partition the dataset into two subsets: an update set, which is used to sequentially train the adapter and measure the update efficacy, and an evaluation set, which is used to measure the forgetting rate of the learned knowledge throughout the sequential updates. 
\end{itemize}

\paragraph{Code Datasets}
\begin{itemize}
    \item \textbf{CodeUpdateArena}~\citep{ref:liu2024codeupdatearena}: A benchmark for knowledge editing in the code domain that evaluates whether language models can internalize API updates. Each instance consists of a synthetic update to an existing API function, paired with multiple program synthesis tasks whose correct solutions require using the updated functionality. The benchmark has 54 functions from 7 Python libraries. To prepare the data, we select one sample per function update for fine-tuning, resulting in a total of 152 training samples, and use the remaining examples for evaluation.
\end{itemize}

\begin{table}[h]
    \centering
    \caption{Statistics of the datasets used for experiments.}
    \label{tab:dataset-stats}
    \begin{tabular}{llcc}
        \toprule
        \textbf{Domain} & \textbf{Dataset} & \textbf{Evaluation} & \textbf{\# Samples}  \\
        \midrule
        Fact & CounterFact & Update Efficacy & 2,191 \\
        & ZsRE & Update Efficacy & 1,000 \\
        & MQuAKE & Update Efficacy & 3,000 \\
        & AToKe-ME & Update Efficacy & 1,230 \\
        & MMLU & Specificity & 14,042 \\
        \midrule
        Code & CodeUpdateArena & Update Efficacy & 670 \\
        & HumanEval & Specificity & 164 \\
        \bottomrule
    \end{tabular}
\end{table}

\subsection{Implementation Details} \label{sec:app-implementation}
\paragraph{LoRA Training.} For training with LoRA, we set the rank to 32 , alpha to 64, and learning rate to $2e-4$ for all factual knowledge update experiments. For code update, we set the rank to 128, alpha to 256, and learning rate to $2e-5$ because code tasks typically involve complex dependencies and logic that require higher model capacity (LoRA rank). Across all experiments, we use a batch size of 4 with 4 gradient accumulation steps.

\paragraph{Baselines.} For F-Learning, we adopt the hyperparameter settings reported in the original paper~\citep{ref:ni2024forgetting}. For MEMIT, we utilize the implementation provided in the EasyEdit framework~\citep{ref:wang2024easyedit}.

\begin{figure}
    \centering
    \includegraphics[width=1.0\linewidth]{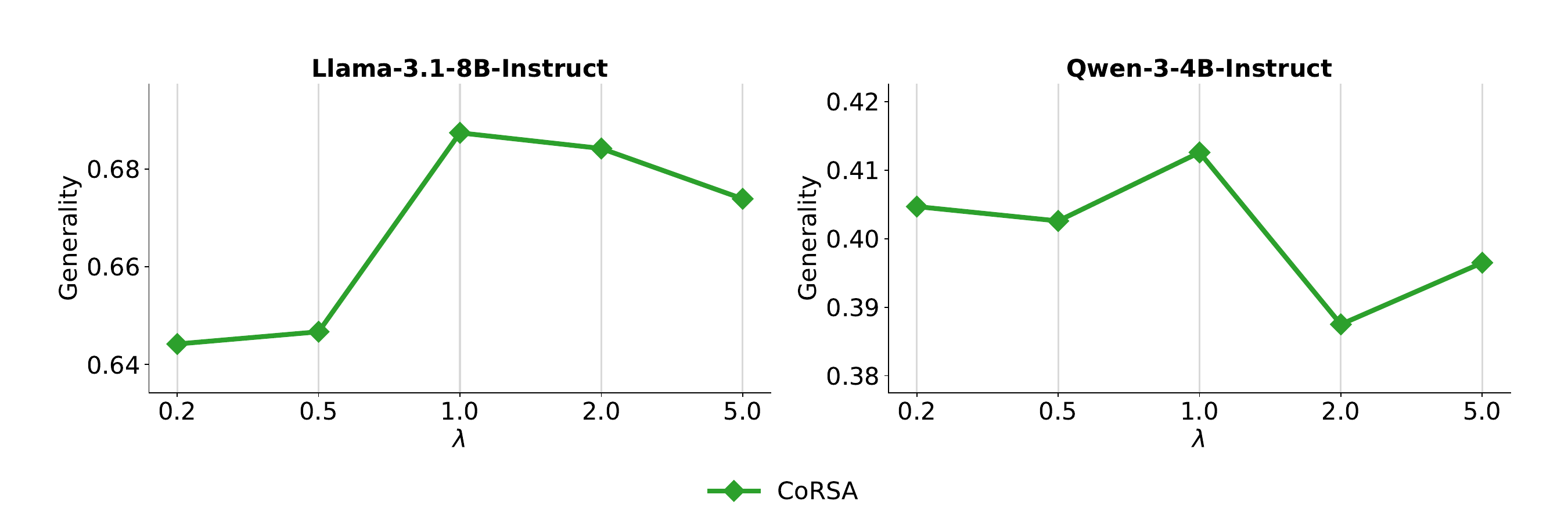}
    \caption{Sensitivity analysis of the hyperparameter $\lambda$ on the CounterFact dataset. We evaluate the Generality score across $\lambda \in \{0.2, 0.5, 1.0, 2.0, 5.0\}$. The results show that performance is maximized at $\lambda=1.0$.} 
    \label{fig:lambda_sensitivity}
\end{figure}

\paragraph{\method{}.} We set $\lambda=1.0$ for every experiment because we aim to assign equal importance to both the SFT objective and the DPO objective for old and new knowledge separation. We further do an analysis on CounterFact to check the sensitivity of our method to this hyperparameter on a development set. As shown in~\cref{fig:lambda_sensitivity}, the Generality consistently peaks at $\lambda=1.0$ for both Llama-3.1-8B-Instruct and Qwen-3-4B-Instruct. This suggests that heavily favoring either objective degrades performance. For SAM, we set $\varepsilon=1e-12$ and perform a hyperparameter sweep for $\rho$ on a small development set in all datasets and observe that $\rho=0.05$ is the best and consistent for training across datasets. For the parameter $\beta$ in the DPO objective, we use the default setting in TRL~\citep{ref:vonwerra2020trl} ($\beta=0.1$) as standard practice to maintain a stable KL-divergence constraint against the reference model.

\subsection{Resources and Training Time} 
\paragraph{GPUs.} Experiments are conducted on four RTX A6000 with 48G memory each.

\paragraph{Training Time.} In our implementation, we observe that \method{} increases training time by approximately $1.8-2.5\times$ compared to standard LoRA. This overhead arises because each optimizer step needs two full forward-backward passes: one to compute the adversarial perturbation and a second to compute the final update gradient. While this theoretically doubles the computational cost ($\approx 2\times$), practical runtime varies due to system overheads such as data loading and DPO-specific computations (e.g., calculating reference log-probabilities). For instance, on the CounterFact dataset and Llama-3.1-8B-Instruct model, a training step takes $\approx$1.3s, whereas a \method{} step requires $\approx$3.1s on a RTX 6000 GPU.

\section{Additional Results} \label{sec:app-results}

\subsection{Adapter Merging}
\begin{figure}
    \centering
    \includegraphics[width=0.5\linewidth]{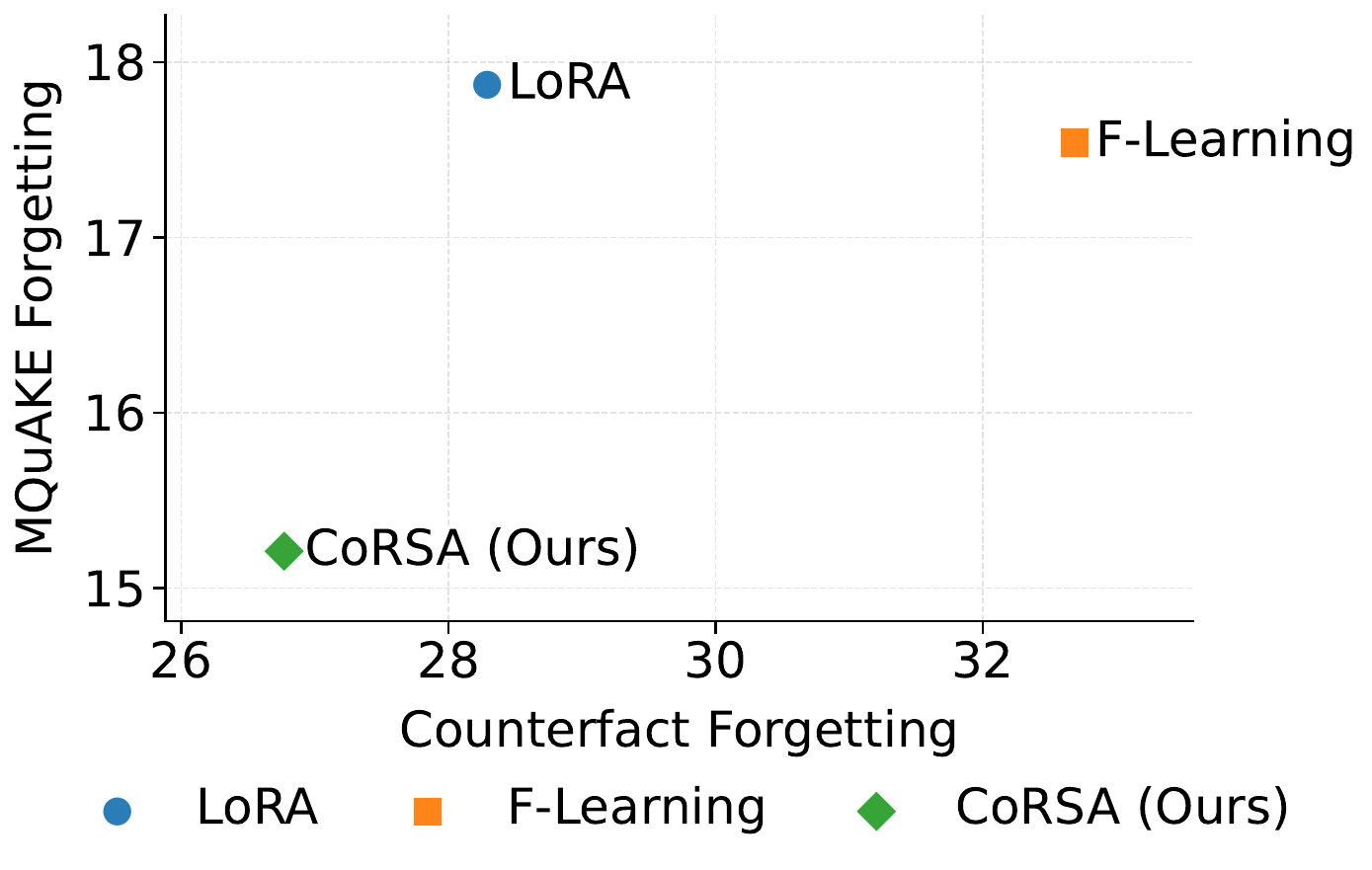}
    \caption{Forgetting rate when merging the adapters trained on CounterFact and MQuAKE datasets using Llama-3.1-8B-Instruct (lower is better for both $x$ and $y$ axes). \method{} achieves the lowest forgetting rate, demonstrating superior ability to retain knowledge from both datasets simultaneously.}
    \label{fig:merging}
\end{figure}

An effective knowledge update framework should also support the composition of distinct knowledge domains through adapter merging without catastrophic forgetting. This capability allows us to combine multiple adapters, each trained on different knowledge sets by merging or composing them to obtain a single adapter that exhibits combined behaviors \citep{ref:ilharco2023editing, ref:yadav2023tiesmerging, ref:zhao2025merging, ref:huang2023lorahub}. In this experiment, we independently train separate LoRA adapters on distinct datasets of knowledge as in~\cref{sec:exp} and subsequently combine them via linear arithmetic merging. We then evaluate the merged model on the union of the respective test sets to determine if the independent updates can coexist without destructive interference (more forgetting). In particular, we evaluate the composability of our method by merging adapters trained on the CounterFact and MQuake datasets.~\cref{fig:merging} show that our methods achieves the lowest forgetting rates across both datasets compared to baselines, demonstrating that our training framework produces more modular adapters that can be merged with minimal interference. Intuitively, merging can be viewed as applying a perturbation to the model weights. Because SAM finds flatter minima that are robust to such perturbations, our adapters maintain high performance when combined.

\subsection{Multiple LoRA Adapters}
While the experiments in~\cref{sec:exp} utilize a single adapter, we further demonstrate that our framework supports multiple adapters. In this setting, we partition the CounterFact dataset into 5 distinct splits, training a LoRA adapter for each knowledge batch. To handle inference, first, we prompt an LLM to generate a high-level natural language description summarizing the specific objects and relations covered in each split. During inference, these summaries serve as semantic description. To route a query, we prompt the LLM to compare the input against these five descriptions and dynamically direct the query to the single most relevant adapter. As shown in Table~\ref{tab:multi-adapter}, our approach achieves an average Generality of 61.94\%, substantially outperforming the standard LoRA baseline (45.36\%) and maintaining performance comparable to the Single Adapter Upper Bound (66.27\%). We also observe that 94.61\% of the gap between our method and the upper bound is caused by routing errors, showing that the adapters themselves remain highly effective. 

\begin{table}[h]
\centering
\small
\caption{Performance evaluation on the CounterFact dataset partitioned into 5 splits, where each split is managed by a distinct LoRA adapter. We report the update efficacy routed via summary-based context for each split and the overall average.}
\label{tab:multi-adapter}
\begin{tabular}{lcccccc}
\toprule
\multirow{2}{*}{\textbf{Method}} & \multicolumn{6}{c}{\textbf{Generality (\%)}} \\
\cmidrule(lr){2-7}
& \textbf{Split 1} & \textbf{Split 2} & \textbf{Split 3} & \textbf{Split 4} & \textbf{Split 5} & \textbf{Average} \\
\midrule
Single Adapter (Upper Bound)
& \textbf{68.82} & \textbf{63.70} & \textbf{67.23} & \textbf{64.15} & \textbf{67.45} & \textbf{66.27} \\
\midrule
LoRA (Summary Routing)
& 45.98 & 42.11 & 46.02 & 46.53 & 46.16 & 45.36 \\
\method{} (Summary Routing)
& \underline{63.54} & \underline{58.51} & \underline{60.97} & \underline{62.87} & \underline{63.81} & \underline{61.94} \\
\bottomrule
\end{tabular}
\end{table}

\subsection{Ablation Study} \label{sec:app-ablation}
To study the effectiveness of each component in \method{}, we conduct an ablation study by systematically removing key elements of our framework, including SAM, DPO, and the conflict-aware gradient projection to isolate their individual contributions to update efficacy. As shown in~\cref{tab:ablation}, removing SAM leads to the greatest drop in Generality on the CounterFact dataset, causing a decline from 66.27\% to 52.78\% on Llama-3.1-8B, which highlights its importance in finding flat minima for better generalization. Similarly, removing DPO substantially decreases performance, particularly on Qwen-3-4B where Generality drops from 39.71\% to 29.03\%. Finally, the removal of PCGrad results in consistent but smaller degradations (e.g., 66.27\% to 63.36\%) on Llama-3.1-8B). Moreover, even with individual components removed, our framework still outperforms LoRA with standard SFT in Generality. These results show that the combination of different components in \method{} yields the highest and robust performance for knowledge update.

\begin{table*}[ht]
\centering
\small
\caption{Ablation study analyzing the contribution of individual components (SAM, DPO, and PCGrad) to the Generality (Gen) and Specificity (Spec) of the model. The full \method{} framework consistently outperforms variants with missing components.
}
\begin{tabular}{lcccccccccccc}
\toprule
& \multicolumn{6}{c}{\textbf{Llama-3.1-8B-Instruct}} & \multicolumn{6}{c}{\textbf{Qwen-3-4B-Instruct}} \\
\cmidrule(lr){2-7} \cmidrule(lr){8-13}
& \multicolumn{2}{c}{CounterFact} & \multicolumn{2}{c}{MQuAKE} & \multicolumn{2}{c}{Avg.}
& \multicolumn{2}{c}{CounterFact} & \multicolumn{2}{c}{MQuAKE} & \multicolumn{2}{c}{Avg.} \\
\cmidrule(lr){2-3} \cmidrule(lr){4-5} \cmidrule(lr){6-7}
\cmidrule(lr){8-9} \cmidrule(lr){10-11} \cmidrule(lr){12-13}
\textbf{Method}
& Gen $\uparrow$ & Spec $\uparrow$ 
& Gen $\uparrow$ & Spec $\uparrow$
& Gen $\uparrow$ & Spec $\uparrow$
& Gen $\uparrow$ & Spec $\uparrow$
& Gen $\uparrow$ & Spec $\uparrow$
& Gen $\uparrow$ & Spec $\uparrow$ \\
\midrule
LoRA (SFT)
& 49.25 & \underline{69.31} 
& 90.97 & 69.89
& 70.11 & 69.60
& 25.29 & 70.47 
& 47.47 & \textbf{71.51}
& 36.38 & \underline{70.99} \\

\method{} 
& \textbf{66.27} & 68.92 
& \textbf{96.53} & \underline{70.17}
& \textbf{81.40} & 69.55
& \textbf{39.71} & \underline{70.81} 
& \textbf{67.17} & \underline{71.28}
& \textbf{53.44} & \textbf{71.05} \\
\midrule
\quad w/o SAM
& 52.78 & \textbf{69.33} 
& 92.23 & \textbf{70.21}
& 72.51 & \textbf{69.77}
& 27.36 & 70.32 
& 49.13 & 70.47
& 38.25 & 70.40 \\

\quad w/o DPO
& 61.07 & 68.98 
& 95.49 & 69.91
& 78.28 & 69.45
& 29.03 & 70.54 
& 54.97 & 70.73
& 42.00 & 70.64 \\

\quad w/o PCGrad
& \underline{63.36} & 69.12 
& \underline{95.97} & 70.16
& \underline{79.67} & \underline{69.64}
& \underline{30.26} & \textbf{70.84} 
& \underline{60.07} & 70.44
& \underline{45.17} & 70.64 \\
\bottomrule
\end{tabular}
\label{tab:ablation}
\end{table*}

\subsection{More Results for~\cref{sec:exp}} \label{sec:app-full-table1}
We show the full results, including the Edit Success metric for the single edit experiment (\cref{tab:fact-editing}) in~\cref{tab:fact-editing-full}. 

\begin{table*}[ht]
\centering
\caption{Full results of single update efficacy across three datasets reporting Edit Success (ES), Generality (Gen), and Specificity (Spec).}
\resizebox{\textwidth}{!}{
\begin{tabular}{lcccccccccccccccccc}
\toprule
& \multicolumn{9}{c}{\textbf{Llama-3.1-8B-Instruct}} & \multicolumn{9}{c}{\textbf{Qwen-3-4B-Instruct}} \\
\cmidrule(lr){2-10} \cmidrule(lr){11-19}
& \multicolumn{3}{c}{CounterFact} & \multicolumn{3}{c}{ZsRE} & \multicolumn{3}{c}{MQuAKE}
& \multicolumn{3}{c}{CounterFact} & \multicolumn{3}{c}{ZsRE} & \multicolumn{3}{c}{MQuAKE} \\
\cmidrule(lr){2-4} \cmidrule(lr){5-7} \cmidrule(lr){8-10}
\cmidrule(lr){11-13} \cmidrule(lr){14-16} \cmidrule(lr){17-19}
\textbf{Method}
& ES & Gen & Spec
& ES & Gen & Spec
& ES & Gen & Spec
& ES & Gen & Spec
& ES & Gen & Spec
& ES & Gen & Spec \\
\midrule
Base
& - & - & 70.28
& - & - & 70.28
& - & - & 70.28
& - & - & 71.74
& - & - & 71.74
& - & - & 71.74 \\

LoRA
& \textbf{98.68} & 49.25 & \textbf{69.31}
& \underline{97.30} & 70.40 & \textbf{70.02}
& \underline{99.47} & 90.97 & \underline{69.89}
& 97.76 & 25.29 & \underline{70.47}
& 99.70 & 68.95 & \underline{71.46}
& 87.47 & 47.47 & \textbf{71.51} \\

MEMIT
& 92.47 & 46.95 & 59.55
& 95.47 & 70.95 & 61.64
& 98.27 & 90.80 & 58.53
& 98.47 & 26.95 & 55.13
& 97.47 & 69.10 & 58.75
& \underline{88.47} & 52.95 & 56.82 \\

F-Learning
& 98.13 & \underline{51.85} & 66.85
& \textbf{98.00} & \underline{71.30} & 67.32
& 99.37 & \underline{95.20} & 65.04
& \textbf{99.04} & \underline{28.71} & 58.29
& \underline{99.78} & \underline{70.00} & 60.17
& 87.69 & \underline{58.70} & 59.42 \\

Ours
& \underline{98.40} & \textbf{66.27} & \underline{68.92}
& \textbf{98.00} & \textbf{72.60} & \underline{69.88}
& \textbf{99.87} & \textbf{96.53} & \textbf{70.17}
& \underline{99.00} & \textbf{39.71} & \textbf{70.81}
& \textbf{99.80} & \textbf{72.10} & \textbf{71.51}
& \textbf{89.54} & \textbf{67.17} & \underline{71.28} \\
\bottomrule
\end{tabular}
}
\label{tab:fact-editing-full}
\end{table*}

\section{Details for \method{}} \label{sec:app-method}
In this section, we provide the technical details for \method{}, including the motivations for SAM (\cref{sec:sam}) and the step-by-step training algorithm (\cref{sec:app-algo}).
\subsection{Sharpness-Aware Minimization (SAM)}
\label{sec:sam}

LLMs or deep neural networks in general often admit many minimizers with similar training loss but very different
local geometry. Solutions that lie in flat regions of the loss landscape
(i.e., regions where the loss does not increase rapidly under small parameter perturbations)
tend to generalize better than sharp solutions. Sharpness-Aware Minimization (SAM)
formalizes this idea by optimizing parameters that perform well not only at a single point
$\theta$, but throughout a neighborhood around $\theta$.

\paragraph{Robust Neighborhood Objective.}
Let $\mathcal{L}_D(\theta)$ denote the training loss on dataset $D$ at parameters $\theta$.
SAM solves the following min--max problem:
\begin{equation}
\label{eq:sam_minmax}
\min_{\theta}\ \max_{\lVert \epsilon \rVert_2 \le \rho}\ \mathcal{L}_D(\theta+\epsilon),
\end{equation}
where $\rho>0$ is the radius of the perturbation ball around the current parameters.
Intuitively, the inner maximization searches for the \emph{worst} nearby parameters
within distance $\rho$, and the outer minimization updates $\theta$ to reduce this worst-case loss.

\paragraph{First-order Approximation of the Inner Maximizer.}
Directly solving the inner maximization in \eqref{eq:sam_minmax} is expensive, so SAM
approximates it via a first-order expansion around $\theta$.
Using $\nabla_{\theta} \mathcal{L}_D(\theta)$ as the gradient at $\theta$, the adversarial perturbation is:
\begin{equation}
\label{eq:sam_eps_star}
\epsilon^\star
:= \arg\max_{\lVert \epsilon \rVert_2 \le \rho} \mathcal{L}_D(\theta+\epsilon)
\ \approx\ 
\rho \cdot \frac{\nabla_{\theta} \mathcal{L}_D(\theta)}{\lVert \nabla_{\theta} \mathcal{L}_D(\theta) \rVert_2}.
\end{equation}
In practice, one often uses a small constant $\varepsilon>0$ for numerical stability,
$\epsilon^\star \approx \rho \cdot \nabla_{\theta} \mathcal{L}_D(\theta)/(\lVert \nabla_{\theta} \mathcal{L}_D(\theta)\rVert_2+\varepsilon)$.

\paragraph{SAM Update Direction.}
After estimating $\epsilon^\star$, SAM evaluates the gradient at the perturbed parameters
and performs a descent step using that gradient:
\begin{equation}
\label{eq:sam_grad}
g^{\text{SAM}}
:= \nabla_{\theta} \left( \max_{\lVert \epsilon \rVert_2 \le \rho}\ \mathcal{L}_D(\theta+\epsilon) \right)
\ \approx\
\nabla_{\theta} \mathcal{L}_D(\theta)\big|_{\theta+\epsilon^\star}
\ =\
\nabla_{\theta} \mathcal{L}_D(\theta+\epsilon^\star).
\end{equation}
Thus, compared to standard ERM (which uses $\nabla_{\theta} \mathcal{L}_D(\theta)$), SAM uses the gradient
at a nearby \emph{adversarially chosen} point to update the original model's parameters.

\paragraph{Why SAM Reduces Sharpness.}
The key mechanism is that SAM explicitly penalizes parameters whose loss
\emph{increases quickly} under small perturbations.
To see the connection to curvature, apply a second-order Taylor expansion:
\begin{equation}
\label{eq:taylor}
\mathcal{L}_D(\theta+\epsilon)
\approx
\mathcal{L}_D(\theta)
+ \nabla_{\theta} \mathcal{L}_D(\theta)^\top \epsilon
+ \tfrac{1}{2}\epsilon^\top H(\theta)\epsilon,
\end{equation}
where $H(\theta):=\nabla_{\theta}^2 \mathcal{L}_D(\theta)$ is the Hessian.
Maximizing \eqref{eq:taylor} over $\lVert \epsilon\rVert_2\le\rho$ yields an upper envelope that depends on both:
(i) the gradient norm (first-order sensitivity) and
(ii) the Hessian spectrum (second-order curvature).
In particular, the quadratic term is controlled by the largest eigenvalue of the Hessian:
\begin{equation}
\label{eq:hessian_bound}
\max_{\lVert \epsilon\rVert_2\le\rho}\ \tfrac{1}{2}\epsilon^\top H(\theta)\epsilon
\ =\
\tfrac{1}{2}\rho^2\,\lambda_{\max}\!\big(H(\theta)\big),
\end{equation}
where $\lambda_{\max}(H)$ is the maximum eigenvalue. More precisely,
$\max_{\lVert \epsilon\rVert_2\le\rho}\epsilon^\top H\epsilon=\rho^2\lambda_{\max}(H)$ for symmetric $H$.
Therefore, minimizing the worst-case neighborhood loss in \eqref{eq:sam_minmax} discourages solutions with large
$\lambda_{\max}(H(\theta))$, i.e., sharp curvature directions that cause the loss to increase rapidly.

\paragraph{Gradient-Level View: Curvature-Aware Descent.}
A complementary perspective comes from expanding the SAM gradient:
\begin{equation}
\label{eq:sam_grad_hessian}
\nabla_{\theta} \mathcal{L}_D(\theta+\epsilon^\star)
\approx
\nabla_{\theta} \mathcal{L}_D(\theta) + H(\theta)\epsilon^\star.
\end{equation}
The additional term $H(\theta)\epsilon^\star$ biases optimization away from directions
where the Hessian amplifies perturbations (high curvature), effectively steering the iterate toward flatter regions.
As a result, SAM tends to find minima that are stable under parameter perturbations, i.e.,
minimizers with reduced local sharpness of the loss landscape.

\paragraph{Formal Connection between Loss Sharpness and Margin Curvature.}
Our update objective is defined as
$\mathcal{L}_{\text{Update}}(\phi) = \mathcal{L}_{\text{SFT}}(\phi) + \lambda \mathcal{L}_{\text{DPO}}(\phi)$,
where SAM directly reduces the local sharpness of the loss. The DPO term can be expressed as a monotonic link function composed with the log-probability margin
$m_\phi(x,y^+,y^-) = \log p_{\theta,\phi}(y^+|x) - \log p_{\theta,\phi}(y^-|x)$.
Specifically, $\mathcal{L}_{\text{DPO}}(\phi) = \mathbb{E}[\ell(\beta m_\phi)]$ where $\ell(z) = -\log\sigma(z)$.
By applying the chain rule, the Hessian of the DPO loss is:
\begin{equation}
    \nabla^2 \mathcal{L}_{\text{DPO}} = \beta\ell'(\beta m_\phi)\nabla^2 m_\phi
    + \beta^2\ell''(\beta m_\phi)\nabla m_\phi (\nabla m_\phi)^\top,
\end{equation}
which implies that the sharpness of $\mathcal{L}_{\text{DPO}}$ depends on both the curvature ($\nabla^2 m_\phi$) and the gradient magnitude ($\nabla m_\phi$) of the margin.
Moreover, since $\nabla^2 \mathcal{L}_{\text{Update}} = \nabla^2 \mathcal{L}_{\text{SFT}} + \lambda\nabla^2 \mathcal{L}_{\text{DPO}}$, the triangle inequality implies:
\begin{equation}
    \|\nabla^2 \mathcal{L}_{\text{DPO}}\|_2 \le \frac{1}{\lambda} \left( \|\nabla^2 \mathcal{L}_{\text{Update}}\|_2 + \|\nabla^2 \mathcal{L}_{\text{SFT}}\|_2 \right).
\end{equation}
Thus, reducing the sharpness of $\mathcal{L}_{\text{Update}}$ via SAM provides an implicit regularization on the DPO term and, indirectly, on margin instability. 

\subsection{Detailed Algorithm for \method{}} \label{sec:app-algo}
In~\cref{alg:method_pc_sam}, we provide the detailed algorithm for training \method{} with SAM and PCGrad. We note that previous work such as~\citet{ref:tran2023sharpness} also explores the gradient conflict problem with SAM, but focuses on memory-replay methods for image classification benchmarks. In contrast, our work addresses the specific conflicts arising in LLM knowledge editing, where the optimization must balance adaptation (SFT) with active suppression (DPO).

\begin{algorithm}[t]
\caption{\method{} Training}
\label{alg:method_pc_sam}
\begin{algorithmic}[1]
\REQUIRE Frozen base model $\theta$, initial LoRA parameters $\phi$. Datasets $\mathcal{D}_{\text{new}}=\{(x_i,y_i^+)\}$, $\mathcal{D}_{\text{pairs}}=\{(x_i,y_i^-,y_i^+)\}$.
\PARAMS Weights $\lambda\ge0$, temperature $\beta>0$, SAM radius $\rho>0$, learning rate $\eta$, constant $\varepsilon>0$.
\ENSURE Updated LoRA parameters $\phi^\star$.

\FOR{$t=1,2,\dots,T$}
    \STATE Sample minibatches $\mathcal{B}_{\text{new}} \sim \mathcal{D}_{\text{new}}$ and $\mathcal{B}_{\text{pairs}} \sim \mathcal{D}_{\text{pairs}}$
    \STATE $g_{\mathrm{SFT}} \gets \nabla_\phi \mathcal{L}_{\mathrm{SFT}}(\phi;\mathcal{B}_{\text{new}}), \quad g_{\mathrm{DPO}} \gets \nabla_\phi \mathcal{L}_{\mathrm{DPO}}(\phi;\mathcal{B}_{\text{pairs}},\theta,\beta)$

    \STATE \textbf{PCGrad (Current Point):} $g_1 \gets g_{\mathrm{SFT}}, \; g_2 \gets g_{\mathrm{DPO}}, \; dp \gets g_1^\top g_2$
    \IF{$dp < 0$} 
        \STATE $g_1 \gets g_1 - \frac{dp}{\|g_2\|_2^2+\varepsilon}\,g_2, \quad g_2 \gets g_2 - \frac{dp}{\|g_1\|_2^2+\varepsilon}\,g_1$
    \ENDIF
    \STATE $g_{\mathrm{PC}} \gets g_1 + \lambda g_2$

    \STATE \textbf{SAM Perturbation:} $\epsilon \gets \rho \frac{g_{\mathrm{PC}}}{\|g_{\mathrm{PC}}\|_2+\varepsilon}, \quad \tilde{\phi} \gets \phi + \epsilon$

    \STATE \textbf{PCGrad (Perturbed Point):} $\tilde{g}_{\mathrm{SFT}} \gets \nabla_\phi \mathcal{L}_{\mathrm{SFT}}(\tilde{\phi};\mathcal{B}_{\text{new}}), \; \tilde{g}_{\mathrm{DPO}} \gets \nabla_\phi \mathcal{L}_{\mathrm{DPO}}(\tilde{\phi};\mathcal{B}_{\text{pairs}},\theta,\beta)$
    \STATE Set $\tilde{g}_1 \gets \tilde{g}_{\mathrm{SFT}}, \; \tilde{g}_2 \gets \tilde{g}_{\mathrm{DPO}}, \; \widetilde{dp} \gets \tilde{g}_1^\top \tilde{g}_2$
    \IF{$\widetilde{dp} < 0$}
        \STATE $\tilde{g}_1 \gets \tilde{g}_1 - \frac{\widetilde{dp}}{\|\tilde{g}_2\|_2^2+\varepsilon}\,\tilde{g}_2, \quad \tilde{g}_2 \gets \tilde{g}_2 - \frac{\widetilde{dp}}{\|\tilde{g}_1\|_2^2+\varepsilon}\,\tilde{g}_1$
    \ENDIF
    \STATE $\tilde{g}_{\mathrm{PC}} \gets \tilde{g}_1 + \lambda \tilde{g}_2$

    \STATE \textbf{SAM Update:} $\phi \gets \phi - \eta\,\tilde{g}_{\mathrm{PC}}$
\ENDFOR
\STATE \textbf{return} $\phi^\star \gets \phi$
\end{algorithmic}
\end{algorithm}

\section{Examples} \label{sec:app-examples}
In this section, we provide the qualitative examples for several results in~\cref{sec:exp}.
Table~\ref{tab:counterfact-examples} shows the qualitative examples of knowledge updating on Llama-3.1-8B and CounterFact.
Table~\ref{tab:mquake-examples} shows examples of multi-hop questions in MQuAKE.
Table~\ref{tab:atoke_reactivation_examples} shows qualitative examples of old-knowledge reactivation on ATOke.

\begin{table*}[t]
\centering
\small
\renewcommand{\arraystretch}{1.2}
\caption{Qualitative examples of knowledge updating on Llama-3.1-8B and CounterFact. Text highlighted in \textcolor{green!60!black}{green} indicates a successful update (matching the target knowledge), while text in \textcolor{red}{red} indicates a failure (hallucination or retention of old knowledge). We also provide two failure cases for every methods at the bottom of the table.}
\label{tab:counterfact-examples}

\begin{tabularx}{\textwidth}{@{} X l l l l @{}}
\toprule
\textbf{Prompt} & \textbf{LoRA} & \textbf{MEMIT} & \textbf{F-Learning} & \textbf{Ours} \\
\midrule

\textbf{Update:} Mount Murphy is located in the continent Antarctica $\rightarrow$ \corr{Africa} \newline
\textbf{Test:} Mount Murphy is a part of the continent of &
\corr{Africa} &
\corr{Africa} &
\corr{Africa} &
\corr{Africa} \\
\addlinespace[0.8em]
\midrule
\textbf{Update:} Bob Mason plays in the position of goaltender $\rightarrow$ \corr{midfielder} \newline
\textbf{Test:} Bob Mason’s playing position is &
\corr{midfielder} &
\corr{midfielder} &
\wrong{composer} &
\corr{midfielder} \\
\addlinespace[0.8em]
\midrule
\textbf{Update:} Mandailing language is located in Indonesia $\rightarrow$ \corr{Poland} \newline
\textbf{Test:} The Mandailing language originates from &
\corr{Poland} &
\wrong{Australia} &
\wrong{India} &
\corr{Poland} \\
\addlinespace[0.8em]
\midrule
\textbf{Update:} Bernard Giraudeau is a native speaker of French $\rightarrow$ \corr{Polish} \newline
\textbf{Test:} The native language of Bernard Giraudeau is &
\wrong{Dutch} &
\wrong{Hebrew} &
\corr{Polish} &
\corr{Polish} \\
\addlinespace[0.8em]
\midrule
\textbf{Update:} 2011 Cannes Film Festival took place in Cannes $\rightarrow$ \corr{Prescott} \newline
\textbf{Test:} Where is 2011 Cannes Film Festival? &
\wrong{Shanghai} &
\wrong{Cannes} &
\wrong{Sweden} &
\corr{Prescott} \\
\addlinespace[0.8em]
\midrule
\textbf{Update:} Space Sentinels premiered on BBC $\rightarrow$ \corr{NBC} \newline
\textbf{Test:} Space Sentinels is to debut on &
\wrong{BBC} &
\wrong{BBC} &
\wrong{CBS} &
\corr{NBC} \\
\addlinespace[0.8em]

\midrule

\textbf{Update:} Charles Lang Freer House, in Detroit $\rightarrow$ \corr{Somerset} \newline
\textbf{Test:} Charles Lang Freer House is located in &
\wrong{Sydney} &
\wrong{England} &
\wrong{Detroit} &
\wrong{Michigan} \\
\addlinespace[0.8em]
\midrule

\textbf{Update:} Edina High School is within Minnesota $\rightarrow$ \corr{Pennsylvania} \newline
\textbf{Test:} Manaudou is a French surname. Edina High School is located in &
\wrong{Ireland} &
\wrong{Rome} &
\wrong{Michigan} &
\wrong{England} \\

\bottomrule
\end{tabularx}
\end{table*}

\begin{table*}[t]
\centering
\small
\setlength{\tabcolsep}{6pt}
\renewcommand{\arraystretch}{1.15}
\caption{Examples of multi-hop questions in MQuAKE.}
\label{tab:mquake-examples}

\begin{tabular}{p{0.47\textwidth} p{0.47\textwidth}}
\toprule
\multicolumn{2}{c}{\large\textbf{Update}} \\
\midrule

\textbf{Update Prompt} &
\textbf{Update Prompt} \\

Cincinnati Gardens is associated with the sport of &
Valerie Hobson is a citizen of \\

\midrule
\textbf{Target} &
\textbf{Target} \\

basketball $\rightarrow$ association football &
United Kingdom $\rightarrow$ Croatia \\

\midrule
\textbf{Multi-hop questions} &
\textbf{Multi-hop questions} \\

In which country was the sport associated with Cincinnati Gardens created? & What is the name of the head of state in the country that Valerie Hobson is a citizen of? \\
\textbf{Answer} & \textbf{Answer} \\
\corr{China} & \corr{Kolinda Grabar-Kitarovi}
 \\

\midrule
\multicolumn{2}{c}{\large\textbf{Generation Output}} \\
\midrule

\textbf{LoRA} &
\textbf{LoRA} \\
Cincinnati Gardens is a venue in the United States of America. The sport associated with Cincinnati Gardens is basketball, but more specifically, it is known for hosting various events including concerts and sports like basketball, but it is most famously associated with the Cincinnati Royals and the Cincinnati Bengals. However, the most well-known sport associated with the venue is basketball. The Cincinnati Royals were a basketball team. The sport of basketball was created in the United States of America by James Naismith. & Valerie Hobson was a British actress. The head of state in the United Kingdom is the monarch, but the head of government is the Prime Minister. However, the question asks for the head of state. The head of state in the UK is the monarch. Valerie Hobson was a British actress, so the country she is a citizen of is the United Kingdom. The current monarch of the UK is King Charles III, but the question does not specify the time period. However, Valerie Hobson was active in the mid-20th century, and during that time, the monarch was Queen Elizabeth II.  \\ Answer: United States of America & Answer: Queen Elizabeth II
 \\

\midrule
\textbf{MEMIT} &
\textbf{MEMIT} \\
The sport associated with Cincinnati Gardens is cricket.
Cricket originated in England. & Valerie Hobson is a citizen of Ireland. Ireland's head of state is the President of Ireland. The President of Ireland is Salva Kiir Mayardit.
 \\
Answer: England & Answer: Salva Kiir Mayardit \\

\midrule
\textbf{F-Learning} &
\textbf{F-Learning} \\
Cincinnati Gardens is described as being in Boynton Beach, Florida. Florida is a state in the United States.
A sport associated with Florida is basketball.
Basketball was created in the United States. & Valerie Hobson was a Croatian, born in 1925. The head of state of Croatia is a monarch, but Croatia is a republic. The head of state of a republic is usually called the President. 
\\
Answer: United States
& Answer: Croatia

 \\

\midrule
\textbf{Ours} &
\textbf{Ours} \\

The sport associated with Cincinnati Gardens is association football. Association football is a sport created in \corr{China}. & Valerie Hobson is a citizen of Croatia. The head of state of Croatia is the President. \corr{Kolinda Grabar-Kitarovi} is the President of Croatia. \\ 
Answer: \corr{China} & Answer: \corr{Kolinda Grabar-Kitarovi} \\

\bottomrule
\end{tabular}
\end{table*}

\section{Prompts} \label{sec:prompts}
We provide the prompts used for fact updating experiments in~\cref{fig:prompt_factual} and code updating experiments in~\cref{fig:prompt_code}.
\begin{figure}[h]
    \centering
    \begin{promptbox}[Prompt for Factual Knowledge Datasets (Direct QA)]
        Q: \{question\} \\
        A:
    \end{promptbox}
    \caption{The prompt template used for factual knowledge datasets.}
    \label{fig:prompt_factual}
\end{figure}

\begin{figure}[h]
    \centering
    \begin{promptbox}[Prompt for Code Datasets]
        [TASK] Your task is to write a Python solution to a problem in a real-world scenario. \\
        Scenario: \\
        \{scenario\} \\
        \\
        Problem: \\
        \{problem\} \\
        \\
        Write a Python function with exactly this solution signature: \\
        \{solution\_signature\} \\
        {}[/TASK] \\
        \\
        {}[TEST] Here are some unit tests to validate your solution: \\
        ```python \\
        \{unit\_tests\} \\
        ``` \\
        {}[/TEST] \\
        \\
        Rules:
        \begin{itemize}
            \item Output ONLY Python code (no markdown).
            \item Define exactly the requested function signature.
        \end{itemize}
    \end{promptbox}
    \caption{The structured prompt template used for code datasets.}
    \label{fig:prompt_code}
\end{figure}

\begin{table*}[t]
\centering
\footnotesize 
\setlength{\tabcolsep}{2pt} 
\renewcommand{\arraystretch}{1.3} %
\newcolumntype{B}{>{\raggedright\arraybackslash\hsize=1.9\hsize}X} 
\newcolumntype{M}{>{\raggedright\arraybackslash\hsize=1.2\hsize}X} 
\newcolumntype{S}{>{\raggedright\arraybackslash\hsize=0.7\hsize}X} 
\newcolumntype{A}{>{\centering\arraybackslash\hsize=0.8\hsize}X} 

\caption{Qualitative examples of old-knowledge reactivation on ATOke.
\textbf{Setting:} We first apply an update to an \emph{updated fact} (Update-1). 
We then apply a \textbf{batch of unrelated updates} (Update-2) and finally re-query the original edited fact. 
Baseline methods often revert to old knowledge (\textcolor{red}{Red}), while \method{} retains the update (\textcolor{green!60!black}{Green}).}
\label{tab:atoke_reactivation_examples}

\begin{tabularx}{\textwidth}{c M B S A A A A}
\toprule
\textbf{\#} &
\textbf{Update-1 \newline (Target)} &
\textbf{Update-2 \newline (Unrelated)} &
\textbf{Re-query \newline (Update-1)} &
\textbf{LoRA} & \textbf{MEMIT} & \textbf{F-Learn} & \textbf{\method{}} \\
\midrule

% Example 1
\textbf{1} &
\textbf{Fact:} Gary Cahill's team is \newline
\textbf{Update:} Chelsea F.C $\rightarrow$ \textbf{Crystal Palace} &
1. Elvedin Beganović's team is FK Sarajevo $\rightarrow$ \textbf{Erzurumspor} \newline
2. Masato Sakurai's team is Kashiwa Reysol $\rightarrow$ \textbf{Tokoha Univ.}
\newline
$\dots$ &
\textbf{Q:} \textit{Gary Cahill plays for} \newline
\textbf{Exp:} \textit{Crystal Palace} &
\textcolor{red}{Chelsea F.C.} &
\textcolor{red}{Arsenal F.C.} &
\textcolor{red}{Nottingham Forest F.C.} &
\textcolor{green!60!black}{Crystal Palace} \\
\midrule

% Example 2
\textbf{2} &
\textbf{Fact:} Álvaro Rivero's team is \newline
\textbf{Edit:} Inter. de Madrid $\rightarrow$ \textbf{Getafe CF B} &
1. Elvedin Beganović's team is FK Sarajevo $\rightarrow$ \textbf{Erzurumspor} \newline
2. Masato Sakurai's team is Kashiwa Reysol $\rightarrow$ \textbf{Tokoha Univ.}
\newline
$\dots$ &
\textbf{Q:} \textit{Álvaro Rivero plays for} \newline
\textbf{Exp:} \textit{Getafe CF B} &
\textcolor{red}{Atlético} &
\textcolor{red}{Internac. de Madrid} &
\textcolor{red}{CD Inter Sevilla} &
\textcolor{green!60!black}{Getafe CF B} \\
\midrule

% Example 3
\textbf{3} &
\textbf{Fact:} Danny Welbeck's team is \newline
\textbf{Edit:} Man. Utd $\rightarrow$ \textbf{Arsenal F.C.} &
1. Elvedin Beganović's team is FK Sarajevo $\rightarrow$ \textbf{Erzurumspor} \newline
2. Masato Sakurai's team is Kashiwa Reysol $\rightarrow$ \textbf{Tokoha Univ.}
\newline
$\dots$ &
\textbf{Q:} \textit{Danny Welbeck plays for} \newline
\textbf{Exp:} \textit{Arsenal F.C.} &
\textcolor{red}{Manchester United} &
\textcolor{red}{Manchester United} &
\textcolor{red}{Manchester United} &
\textcolor{green!60!black}{Arsenal F.C.} \\
\bottomrule
\end{tabularx}
\end{table*}

\end{document}